\documentclass[a4paper,fleqn]{cas-dc}
\usepackage[numbers,sort&compress]{natbib}
\usepackage{microtype}
\graphicspath{{01_current_appendix/}{02_candidates/}{./}{figs/}{els-cas-templates/}{els-cas-templates/figs/}}
\hypersetup{colorlinks=true,linkcolor=black,citecolor=black,urlcolor=black}
\newcommand{\bestcell}[1]{\textcolor[rgb]{0.65,0.12,0.12}{\textbf{#1}}}
\newcommand{\secondcell}[1]{\textcolor[rgb]{0.10,0.29,0.55}{#1}}

\renewcommand{\printorcid}{}
\ExplSyntaxOn
\keys_set:nn { stm / mktitle } { nologo }
\ExplSyntaxOff

\begin{document}
\let\WriteBookmarks\relax
\raggedbottom
\setcounter{topnumber}{1}
\setcounter{bottomnumber}{1}
\setcounter{totalnumber}{2}
\setcounter{dbltopnumber}{1}
\renewcommand{\topfraction}{0.76}
\renewcommand{\dbltopfraction}{0.76}
\renewcommand{\textfraction}{0.18}
\renewcommand{\floatpagefraction}{0.75}
\renewcommand{\dblfloatpagefraction}{0.75}
\setlength{\textfloatsep}{14pt plus 3pt minus 2pt}
\setlength{\dbltextfloatsep}{15pt plus 3pt minus 2pt}
\setlength{\floatsep}{12pt plus 2pt minus 2pt}
\shorttitle{Target-relative spectral representations for BOLD prediction}
\shortauthors{Liu et al.}
\title[mode=title]{MEL: Target-relative spectral representations for EEG-based regional BOLD prediction}

\author[1]{Xiangyu Liu}
\fnmark[1]
\ead{t330034034@mail.bnbu.edu.cn}
\author[2]{Zeting Yan}
\fnmark[1]
\ead{zetingyan@connect.hku.hk}
\author[3]{Zhitong Yin}
\ead{2300012924@stu.pku.edu.cn}
\author[3]{Boyang Li}
\cormark[1]
\ead{2101112018@stu.pku.edu.cn}
\author[3]{Xi Zhang}
\cormark[1]
\ead{xi.zhang@pku.edu.cn}
\affiliation[1]{organization={Beijing Normal-Hong Kong Baptist University},city={Zhuhai},country={China}}
\affiliation[2]{organization={The University of Hong Kong},city={Hong Kong},country={China}}
\affiliation[3]{organization={Peking University},city={Beijing},country={China}}
\fntext[1]{These authors contributed equally to this work.}
\cortext[1]{Corresponding authors.}

\begin{highlights}
\item Target-relative EEG coordinates support controlled spectral comparisons.
\item Morlet features improve participant-paired prediction with ridge regression.
\item The estimator advantage does not require additional nonlinear readout capacity.
\item Refitted controls separate feature identity from arbitrary coordinate order.
\item Participant-separated and external analyses delimit predictive generalization.
\end{highlights}

\begin{abstract}
\textbf{Background and Objective:} Predicting regional blood-oxygen-level-dependent (BOLD) activity from EEG provides a computational approach to studying electrophysiological--hemodynamic associations. However, changes in input representation, readout architecture, and ensembling can obscure the source of a prediction gain. We investigate the contribution of target-relative spectral estimation under matched readout conditions.

\textbf{Methods:} MEL converts each preceding 16-second EEG history into calibrated lag--channel--frequency measurements. Four equal-dimensional representations are compared using matched Ridge and multilayer perceptron (MLP) readouts. The retrospective analysis includes 29 resting-state scans from 22 participants, participant-paired uncertainty estimates, aggregation and correspondence controls, leave-one-participant-out regression, and refitted models on an external Oddball cohort.

\textbf{Results:} Morlet--Ridge achieved a mean scan-wise correlation of 0.508 across six regional targets and the global signal, compared with 0.432 for canonical Fourier features. The participant-paired mean difference was 0.069 (95\% bootstrap confidence interval, 0.037--0.101), with improvements in 19 of 22 participants. The Morlet MLP prediction ensemble reached 0.491. Shared coordinate permutations preserved prediction after refitting, whereas sample-wise permutations degraded it. Canonical Fourier regression reached 0.377 under participant exclusion; external unseen-participant correlation was 0.026.

\textbf{Conclusions:} Spectral estimation materially affected regional BOLD prediction without requiring additional nonlinear readout capacity. MEL provides an explicit measurement interface for controlled comparisons, but weak external results constrain its generalizability. Confirmation requires untouched participant-separated data and split-aware preprocessing.
\end{abstract}
\begin{keywords}
EEG--fMRI \sep BOLD prediction \sep Spectral representation \sep Morlet wavelets \sep Regularized regression \sep Reproducibility
\end{keywords}
\maketitle

\section{Introduction}
Relating scalp electrophysiology to regional hemodynamic activity is a central challenge in multimodal brain research. Electroencephalography (EEG) resolves rapid electrical fluctuations, whereas functional magnetic resonance imaging (fMRI) measures spatially distributed blood-oxygen-level-dependent (BOLD) signals. Simultaneous EEG--fMRI enables their joint study but requires scanner access and specialized acquisition procedures \cite{huster2012simultaneous,abreu2018eeginformed}. Predicting regional BOLD time courses from EEG offers a computational means of examining their shared information. Establishing when such prediction is reliable is also a prerequisite for considering EEG-based regional monitoring beyond simultaneous acquisition.

The EEG evidence relevant to a BOLD target is distributed across time and frequency. Neurovascular dynamics relate BOLD measurements to preceding activity rather than to a single simultaneous electrical observation \cite{logothetis2001fmri,buxton1998balloon}, and EEG rhythms exhibit region- and state-dependent associations with hemodynamic fluctuations \cite{mantini2007signatures,scheeringa2011neuronal}. Representing the preceding history therefore entails consequential measurement choices. A whole-history spectrum discards the temporal distribution of spectral activity; interval-based measurements retain that distribution but depend on the spectral estimator and its temporal support. These choices determine which distinctions remain available to a downstream predictor.

\begin{figure}[pos=tb]
\centering
\includegraphics[width=\columnwidth]{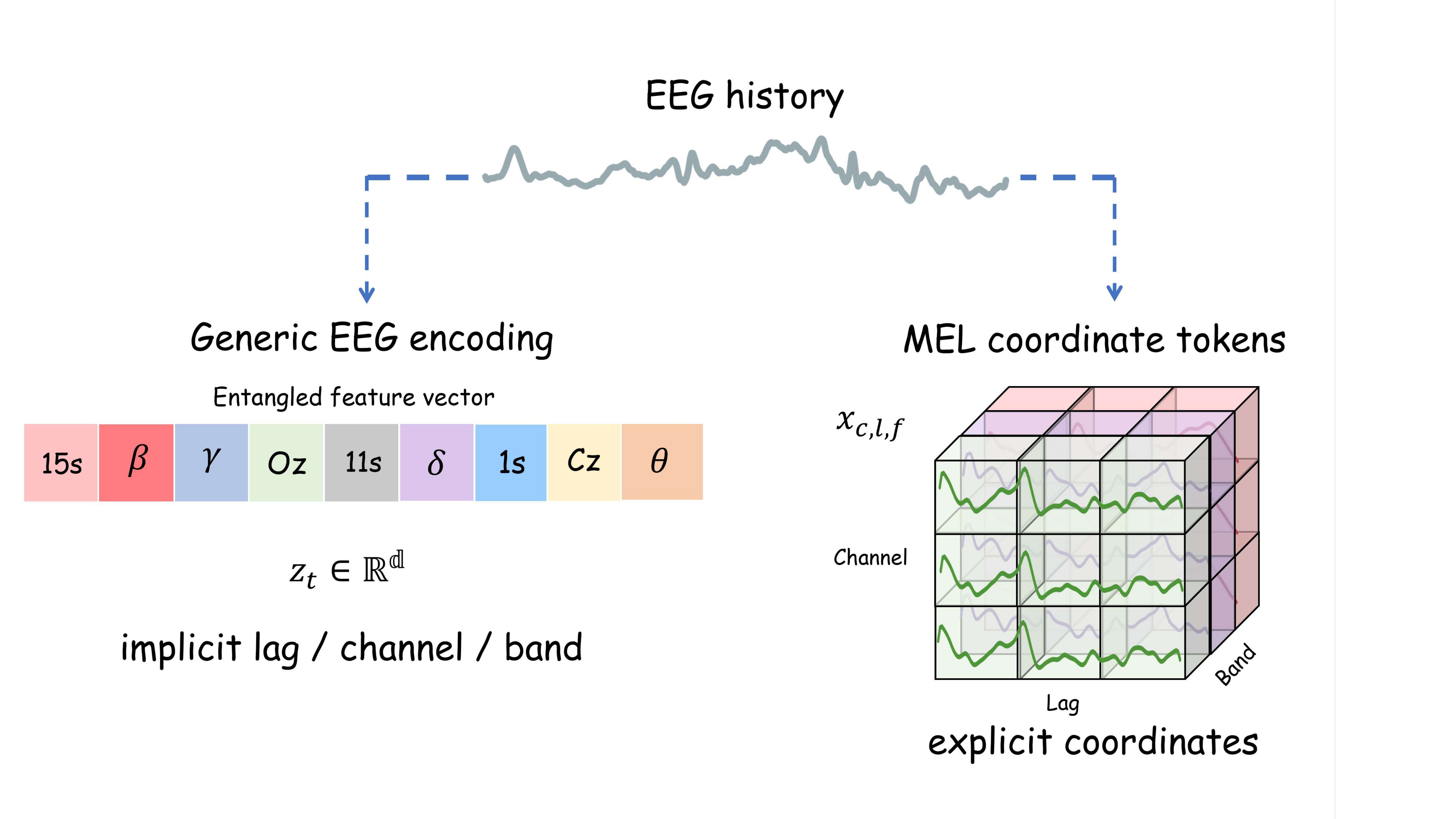}
\caption{Motivation for retaining identifiable lag, channel, and frequency measurements before regional BOLD prediction. Encoder drawings are schematic; learned representations can also encode positional information.}
\label{fig:motivation}
\end{figure}

Regional time-course prediction provides a focused endpoint for investigating these choices. Unlike whole-volume synthesis, it evaluates temporal variation in a specified set of outputs and permits direct inspection of target-wise errors. Nevertheless, an aggregate correlation cannot establish amplitude fidelity, uniform regional performance, or generalization to new participants. A representation study must therefore connect its measurement design to paired prediction evidence, regional heterogeneity, and clearly distinguished evaluation settings. This requirement motivates treating the EEG measurement interface as an experimental variable rather than a fixed precursor to network design.

EEG-to-fMRI methods have investigated electrode relationships, multidimensional feature learning, transformers, and convolutional encoder--decoders \cite{calhas2023attentional,li2024neurobolt,lanzino2024ntvit,roos2025e2fnet}. These developments expand the models available for cross-modal prediction. However, comparisons that change the input representation, readout architecture, and ensembling procedure simultaneously cannot attribute an observed gain to any one component. A complementary question is whether the measurements themselves can improve prediction when the downstream readout is held fixed.

A deterministic spectral interface makes this question experimentally tractable. Established Fourier and wavelet operations produce explicitly defined measurements that can be exchanged, aggregated, or perturbed without changing the prediction target. Learned encoders remain valuable for discovering task-adapted interactions; the purpose of an explicit interface is not to replace them, but to establish a controlled reference against which their additional contribution can be assessed.

Two distinctions are essential to this analysis. First, equal-dimensional representations can differ in bandwidth, taper, and temporal support: matching feature counts controls input size, not measurement equivalence or effective model complexity. Second, organizing features as a lag--channel--frequency tensor does not imply that a predictor operating on its flattened form exploits tensor geometry. Matched readouts address the first attribution problem, while aggregation and correspondence controls examine the information retained by the coordinate definitions. Figure~\ref{fig:motivation} illustrates the underlying measurement-level rationale.

We develop and evaluate MEL (Multi-band EEG Latent-state Tokenization), an explicit target-relative interface for regional BOLD prediction. Each pre-target EEG history is converted into a calibrated lag--channel--frequency tensor whose measurements remain identifiable after vectorization. Fourier and Morlet estimators share this interface, allowing spectral measurement and readout design to be varied separately. Here, tokenization denotes the organization of deterministic spectral measurements; no physiological latent-state model is fitted.

\begin{figure*}[pos=t]
\centering
\includegraphics[width=\textwidth]{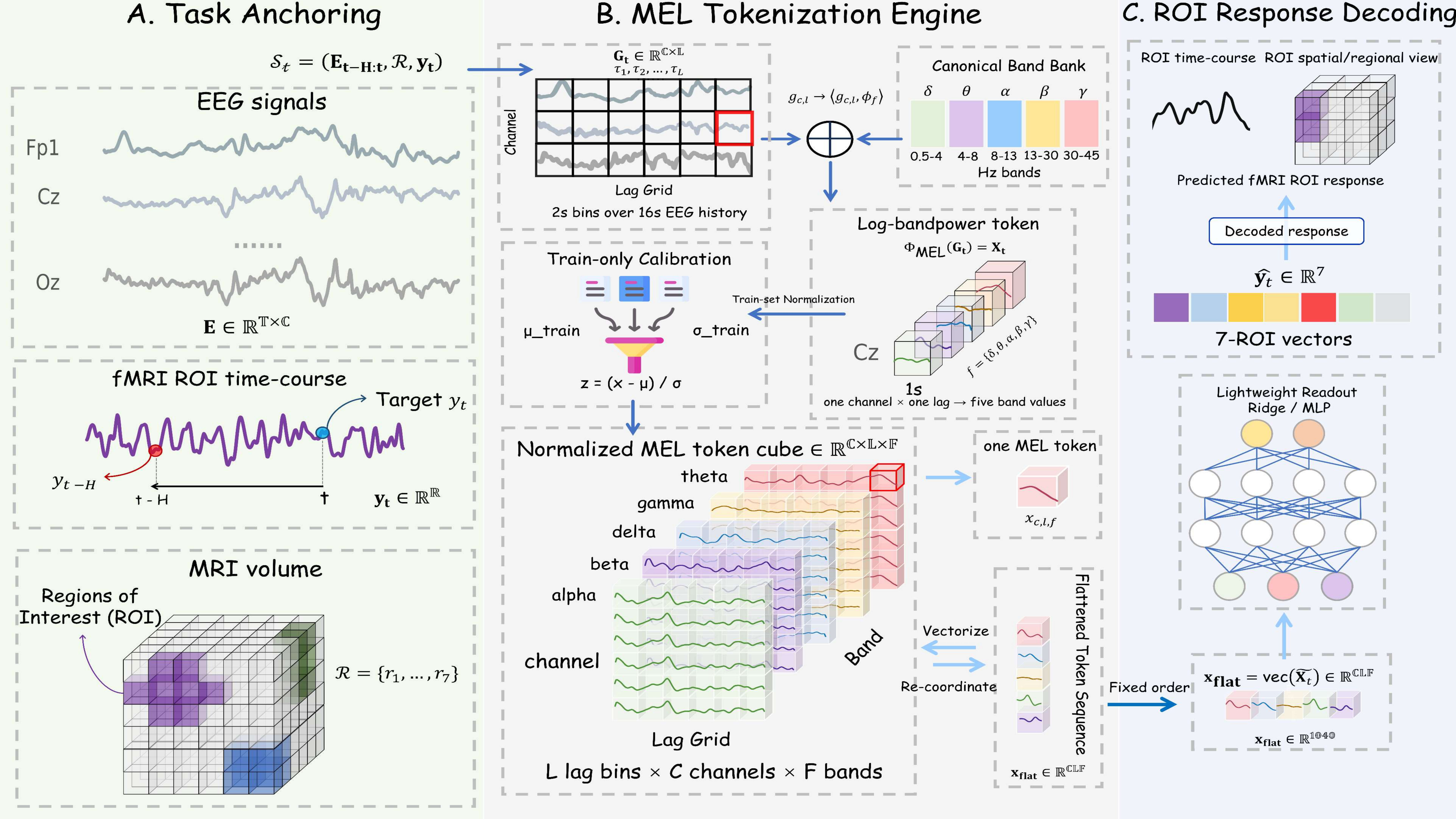}
\caption{MEL pipeline illustrated with canonical Fourier measurements. Eight two-second intervals yield 1040 calibrated coordinates for Ridge or MLP prediction of six regional BOLD time courses and the global signal. BOLD responses provide supervision, not prediction inputs.}
\label{fig:framework}
\end{figure*}

The primary objective is to determine whether spectral estimation improves prediction without requiring a more expressive readout. Four equal-dimensional representations are evaluated within matched Ridge and MLP families, with participant-paired and target-wise analyses. Aggregation and refitted correspondence controls then examine which distinctions support prediction and whether their storage order matters. Finally, participant-exclusion and external analyses assess the portability of canonical Fourier measurements. The endpoint throughout is regional time-course prediction, comprising six anatomical targets and the global signal in the primary cohort.

The contribution is a fully specified measurement interface together with a controlled analysis of its predictive behavior. Rather than introducing a new spectral transform, the study separates estimator effects from readout and ensemble effects, distinguishes coordinate identity from arbitrary storage order, and evaluates prediction under explicitly different generalization protocols. This sequence connects a design choice to its measured benefit, its interpretation, and its limits. Numerical and statistical specifications are provided in \ref{app:implementation} through \ref{app:generalization}.

\section{Methods}
MEL comprises three stages: target-relative history construction, spectral measurement with training-derived calibration, and regional prediction (Figure~\ref{fig:framework}). This separation permits changes to the measurements to be evaluated within a fixed readout family. BOLD time courses serve as responses for fitting and evaluation and are never supplied as readout inputs.

\subsection{Study objectives and comparison strategy}
The analysis follows three questions. First, how does spectral estimation affect prediction when input dimension, readout architecture, and selection budget are held fixed? Second, how does prediction change when coordinate distinctions are aggregated or their identities are disrupted? Third, how well do canonical Fourier measurements support prediction when the evaluated participant is excluded from calibration and fitting? External evaluation extends the last question to a different acquisition and task setting.

The primary comparison crosses four 1040-dimensional representations with Ridge, a fixed MLP, and a prediction ensemble of the same five MLPs. Representations within each family share histories, partitions, architecture, and selection budget. This design controls the readout specification rather than equating the effective capacity or optimization of Ridge and MLP. Aggregation experiments address a different question: they change both the retained information and the geometry of the regularized fit. Their effects are therefore interpreted as comparisons between complete representations, not as isolated causal contributions of individual axes. The protocol ledger in \ref{app:selection} distinguishes the analysis families.

\subsection{Data and preprocessing}
\subsubsection{Resting-state cohort and prediction targets}
The primary dataset is the resting-state EEG--fMRI cohort released with NeuroBOLT \cite{li2024neurobolt}. The analysis includes 29 scans from 22 participants, with 26 retained EEG channels sampled at 200 Hz and an fMRI repetition time of 2.1 seconds. Six regional targets are defined by the supplied DiFuMo-based processing: the cuneus, Heschl's gyrus, anterior middle frontal gyrus, anterior precuneus, putamen, and thalamus. The seventh output is the global BOLD signal, which is an aggregate target rather than an anatomical region of interest (ROI).

For reporting, the regional targets are grouped as sensory, higher-level cognitive, and subcortical, with the global signal shown separately. These groups organize regional heterogeneity rather than define mechanistically homogeneous classes. The aggregate endpoint weights each of the seven outputs equally, so the global signal contributes the same numerical weight as any one regional target. Participants, rather than targets, define the independent units used for paired uncertainty estimation.

\subsubsection{EEG--BOLD alignment and chronological partitions}
Each BOLD target is paired with the preceding 16-second EEG history. Chronological partitioning yields 453 training, 46 validation, and 48 test examples per scan, with ten repetition times (21 seconds) reserved at each boundary. Auxiliary EOG, EMG, ECG, and carbon-wire-loop channels are excluded. All intra-scan representations use the same aligned histories, responses, and partitions, confining the comparison to downstream feature construction and prediction.

Target-relative alignment fixes the reference point before feature construction: the most recent interval ends at the target boundary, and earlier intervals represent progressively older observations. The readout receives EEG-derived measurements rather than a BOLD history. Because the 16-second window exceeds the 2.1-second target spacing, consecutive examples share substantial EEG support. Boundary guards separate the histories assigned to different partitions, but observations within a test partition remain temporally dependent.

The inherited pipeline applies fifth-order 0.15-Hz forward--backward BOLD filtering and whole-scan target normalization before partitioning. These operations introduce cross-partition dependencies in the response processing, making the analysis retrospective. They are retained to maintain a common preprocessing basis across representations; training-only feature calibration does not eliminate their effects. All subsequent feature-calibration and model-fitting procedures use the designated training partitions, with model selection restricted to validation data.

\begin{figure*}[pos=t]
\centering
\includegraphics[width=\textwidth]{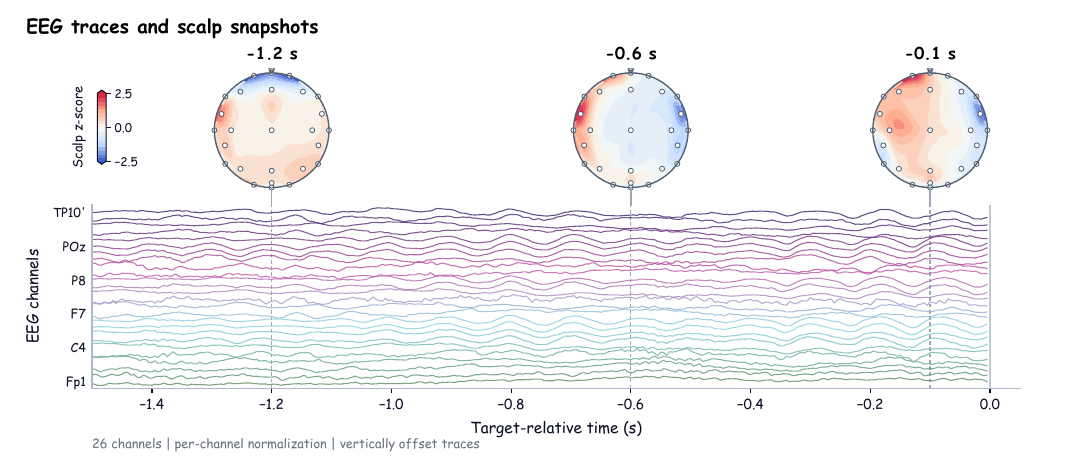}
\caption{Illustrative EEG history from participant 01, scan 01. The 26 traces span $-1.5$ to 0 s and are standardized within the displayed crop, with vertical offsets. Scalp snapshots show cross-channel-standardized means over approximately $\pm0.12$ s around the indicated times.}
\label{fig:eeg_history}
\end{figure*}

Figure~\ref{fig:eeg_history} illustrates the short-timescale variation within an observed pre-target history. The traces and scalp snapshots are descriptive views of the input, not evidence of a fitted EEG--BOLD mechanism.

\subsection{Target-relative spectral representation}
For a BOLD target at time $t_n$, let $X_n\in\mathbb{R}^{C\times Hf_s}$ denote its preceding EEG history, with $H=16$ seconds, $C=26$, and $f_s=200$ Hz. Eight nonoverlapping intervals of duration $\Delta=2$ seconds partition $[t_n-H,t_n)$ chronologically. The segmented history has shape $L\times C\times S$, where $L=8$ and $S=\Delta f_s=400$. Spectral estimation and aggregation replace the within-interval sample axis with $F=5$ measurements, producing $G_n\in\mathbb{R}^{L\times C\times F}$.

Each token is a scalar measurement identified by a temporal interval, recording channel, and spectral definition. No learned embedding or quantization is introduced. The resulting coordinates can be retained, pooled, or recomputed with another estimator while preserving target alignment. Here, \emph{coordinate-preserving} denotes the ability to trace a vectorized feature back to its measurement definition; it does not imply lossless EEG reconstruction or a topology-constrained readout.

\subsubsection{Fourier measurements}
The canonical estimator applies a Hann window and a real discrete Fourier transform to each two-second interval. Squared Fourier magnitudes are averaged over $[0.5,4)$, $[4,8)$, $[8,13)$, $[13,30)$, and $[30,45)$ Hz, then compressed using $\log(1+p)$. These features are band-averaged squared magnitudes, not density-normalized physical power estimates. Uniform Fourier features instead use five equally spaced frequency intervals from 0.5 to 45 Hz. The whole-history comparator pools a single 16-second spectrum into 40 frequency intervals per channel, exchanging an explicit lag axis for finer spectral subdivision. All four primary representations contain 1040 features.

\subsubsection{Morlet energy measurements}
The alternative estimator uses complex Morlet convolution, an established time--frequency analysis method \cite{cohen2014neural}. Each channel history $x_{n,c}\in\mathbb{R}^{Hf_s}$ is convolved with kernels centered at 2, 6, 10.5, 21.5, and 37.5 Hz, using 3, 4, 5, 7, and 9 cycles, respectively. Gaussian envelopes are truncated at $\pm3.5$ temporal standard deviations; the discrete kernels are made zero-mean and normalized to unit energy. Zero-padded FFT convolution followed by centered cropping preserves the history length. Energy is averaged within each of the eight intervals:
\begin{equation}
g_{n,l,c,f}=\log\!\left(1+\frac{1}{S}\sum_{s\in\mathcal I_l}
\left|(x_{n,c}*\psi_f)[s]\right|^2\right).
\label{eq:morlet}
\end{equation}
Here, $\mathcal I_l$ indexes the $S$ samples in interval $l$, and $*$ denotes discrete convolution. The resulting coordinates have frequency-dependent bandwidths rather than the rectangular passbands used for Fourier averaging. Convolution is restricted to the available pre-target history and does not use post-target EEG.

The Fourier and Morlet paths share an output layout but differ in measurement support. Fourier features are computed from independently windowed two-second segments; Morlet energy is estimated over the complete history before interval averaging, allowing measurements near an internal boundary to depend on neighboring intervals. The comparison therefore evaluates a complete estimator choice, including bandwidth, taper, and temporal support, rather than isolating a single kernel property. Boundary handling and the discrete kernels are specified in \ref{app:implementation}.

\begin{figure*}[pos=t]
\centering
\begin{minipage}[t]{0.48\textwidth}
\vspace{0pt}
\centering
\includegraphics[width=\linewidth]{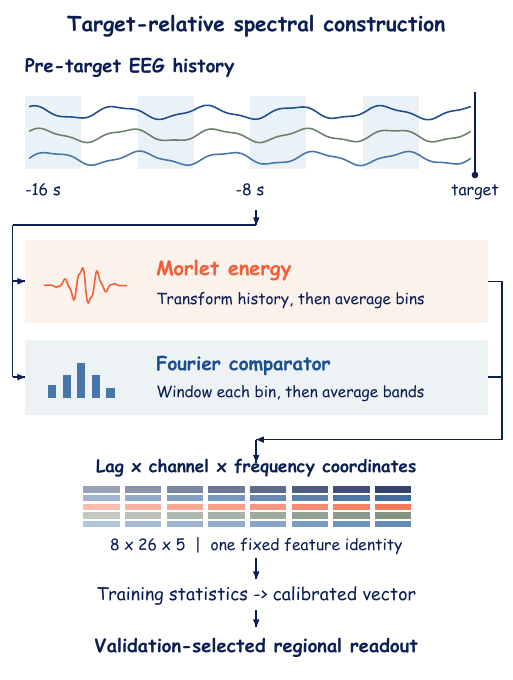}
\end{minipage}\hfill
\begin{minipage}[t]{0.48\textwidth}
\vspace{0pt}
\centering
\includegraphics[width=\linewidth]{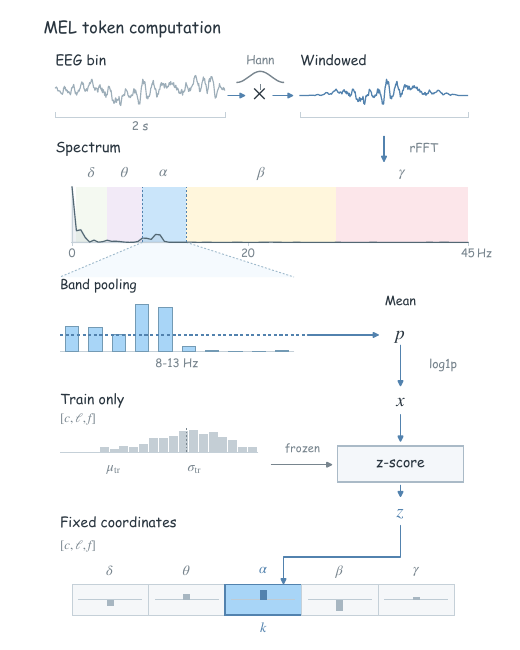}
\end{minipage}
\caption{Spectral coordinate construction. Left: Morlet and Fourier estimators share the output tensor shape. Right: windowing, spectral pooling, log compression, and training-only calibration for one Fourier coordinate. Waveforms and tiles are schematic.}
\label{fig:method}
\end{figure*}

Figure~\ref{fig:method} illustrates both estimator paths and the construction of a single Fourier coordinate.

\subsubsection{Training-derived calibration and vectorization}
Each coordinate is standardized using the mean and standard deviation estimated from training examples; a standard deviation below $10^{-6}$ is replaced by one. The calibrated tensor $Z_n\in\mathbb{R}^{L\times C\times F}$ is vectorized with zero-based index $j=(lC+c)F+f$, yielding $z_n\in\mathbb{R}^{D}$ with $D=1040$. Any additional scaling after aggregation or projection is also fitted exclusively to training data.

Calibration places differently scaled measurements on a common footing for regularized fitting without using validation or test inputs to estimate feature statistics. Vectorization changes their storage layout while preserving the correspondence between each coefficient and its lag, channel, and spectral measurement. This correspondence supports inspection of the fitted model, but correlated predictors prevent coefficient magnitude from being interpreted as an independent physiological contribution. Descriptive scalp and spectral displays use their own stated normalization rules and are not substituted for the model inputs.

\begin{figure*}[pos=t]
\centering
\includegraphics[width=\textwidth]{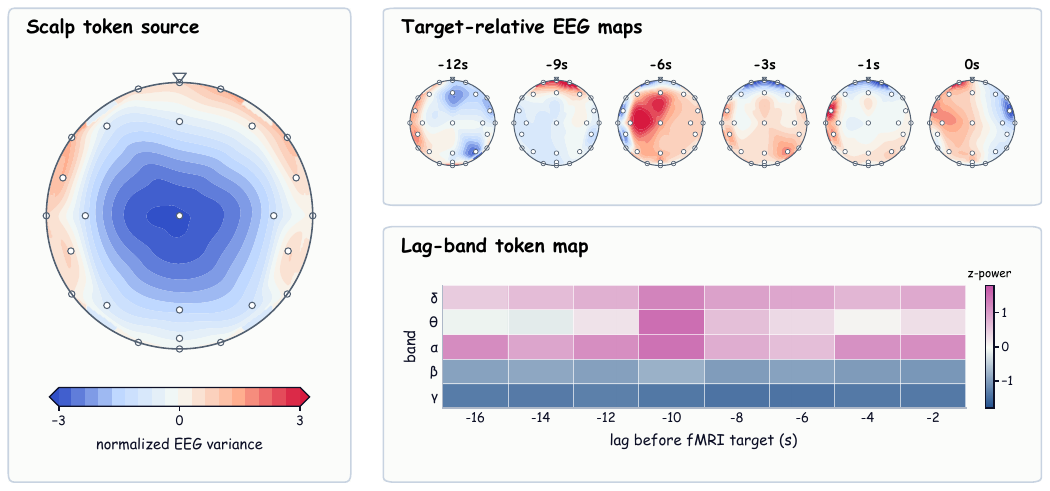}
\caption{Descriptive EEG coordinates. Left: cross-channel-standardized log variance at $-10$ to $-8$ s. Upper right: similarly standardized local means. Lower right: channel-pooled Welch power with display-specific normalization and delta bounds (1--4 Hz), not the prediction input.}
\label{fig:coordinate_display}
\end{figure*}

Figure~\ref{fig:coordinate_display} links the coordinate definitions to one observed EEG history; display normalization remains separate from prediction calibration.

\subsection{Prediction models and model selection}
\subsubsection{Ridge and matched nonlinear readouts}
For target $r$ and training index set $\mathcal T$, ridge regression \cite{hoerl1970ridge} minimizes
\begin{equation}
\sum_{n\in\mathcal T}(y_{n,r}-b_r-z_n^\top w_r)^2
+\lambda_r\|w_r\|_2^2,
\label{eq:ridge}
\end{equation}
with an unpenalized intercept. Each target selects its penalty by maximizing validation Pearson correlation. The matched-readout experiment evaluates 33 logarithmically spaced penalties from $10^{-1}$ to $10^7$. A separate estimator-and-aggregation analysis uses 31 penalties from $10^{-2}$ to $10^8$ with an LSQR solver. Results from these configurations are distinguished throughout because they reuse the cohort and do not constitute independent replications.

The matched MLP consists of input layer normalization, a 16-unit fully connected layer, GELU, dropout of 0.1, and a seven-output linear layer, totaling 18,855 trainable parameters. Each representation uses the same architecture, five fixed seeds, AdamW with learning rate $10^{-3}$ and weight decay $10^{-3}$, batch size 128, and mean squared error (MSE) loss. Training is limited to 60 epochs, with early-stopping patience 12 and gradient clipping at one. Each seed retains the checkpoint with the highest validation mean correlation. Test performance is used to select neither the checkpoint nor the seed.

Two MLP endpoints are reported. Averaging the five seed-specific correlations summarizes the single-model training procedure over random initializations; averaging the five prediction vectors before scoring evaluates a multi-model predictor. Because correlation is nonlinear, these operations are not interchangeable. Reporting both endpoints separates the estimator contrast from the benefit of ensembling and retains every fixed seed rather than selecting the best test result.

\subsubsection{Computational requirements}
For $N$ histories, Fourier construction has approximate cost $O(NCLS\log S)$, while FFT-based Morlet construction costs $O(NCFM\log M)$ for padded convolution length $M$. Calibration costs $O(ND)$. Linear inference requires $O(DK)$ operations per target time for $K=7$ outputs and stores $DK$ coefficients and $K$ intercepts (7280 and seven, respectively). A dense Ridge fit can be formulated in the primal, $O(ND^2+D^3)$, or dual, $O(N^2D+N^3)$, per target and penalty; LSQR instead uses iterative matrix products. Validation searches multiply the fitting workload. These analytical costs characterize the operations but do not establish measured end-to-end efficiency.

\subsection{Controlled representation experiments}
\subsubsection{Axis aggregation and compact representations}
The estimator-and-aggregation analysis compares full Morlet coordinates with channel-, frequency-, and lag-pooled variants, concatenated axis marginals, and channel means and standard deviations at each lag--frequency location. Principal component analysis (PCA) and partial least squares (PLS), both fitted to training data, provide compact comparators. Prediction fusion uses validation-selected combination weights. Together, these comparisons examine the utility and redundancy of the retained distinctions.

Pooling removes distinctions along one axis while retaining the other two; concatenated marginals retain several lower-order summaries without the full joint array. PCA evaluates unsupervised compression, whereas PLS uses training responses to construct supervised projections. Fusion tests the complementarity of prediction vectors. Feature dimensions and selection procedures are specified in \ref{app:selection}, allowing each outcome to be interpreted in relation to both information reduction and its associated fitting procedure.

\subsubsection{Coordinate and feature--target correspondence}
Vectorization preserves measurement identities without imposing tensor geometry on Ridge or an unconstrained MLP. For Ridge, a shared permutation $P$ can be absorbed into the coefficients because $z^\top w=(Pz)^\top(Pw)$ and the penalty is invariant. We therefore distinguish consistent reindexing from disrupted correspondence. A shared channel permutation, lag reversal, band reversal, or flat permutation is applied across all partitions before refitting. Other controls independently permute each example, circularly shift features relative to targets within partitions, or shuffle training labels. Every readout is re-estimated. The contrast tests whether prediction depends on a stable measurement-to-response mapping rather than on a privileged storage order.

\subsubsection{Temporal resolution and cached support}
Additional canonical Fourier diagnostics vary training fraction, extra partition guards (0--24 target times), and bin duration (one to eight seconds). Training subsets are sampled independently for each representation, so these curves do not define paired sample-efficiency comparisons. The bin-duration analysis uses the historical shared-penalty procedure; the other diagnostics use target-specific penalties. All retain the cached preprocessing and are reported separately from the primary matched-readout analysis.

\subsection{Participant-separated and external evaluation}
\subsubsection{Leave-one-participant-out regression}
Leave-one-participant-out regression excludes every scan of the evaluated participant. For canonical Fourier measurements, uncalibrated log-bandpower features from the remaining participants' training partitions are used to fit the scaler and readout; their validation partitions select target-specific penalties from 31 values between $10^{-2}$ and $10^8$. Evaluation is restricted to the held-out participant's test intervals. Thus, that participant contributes neither feature-calibration statistics nor fitting observations. The comparison includes lagged, time-averaged, and fixed temporally weighted Fourier features; it does not evaluate Morlet features or MLPs across participants.

% BEGIN EMBEDDED table_main_results.tex
\begin{table*}[pos=t]
\centering
\scriptsize
\setlength{\tabcolsep}{2.7pt}
\renewcommand{\arraystretch}{1.12}
\resizebox{\textwidth}{!}{
\begin{tabular}{llcccccccc}
\toprule
\textbf{Setting} & \textbf{Representation} & \multicolumn{2}{c}{\textbf{Primary Sensory}} & \multicolumn{2}{c}{\textbf{High-level Cognitive}} & \multicolumn{2}{c}{\textbf{Subcortical}} & \textbf{Global Signal} & \textbf{Avg. $R\uparrow$} \\
\cmidrule(lr){3-4}\cmidrule(lr){5-6}\cmidrule(lr){7-8}
& & Cuneus & Heschl's Gyrus & \makecell{Middle\\Frontal} & \makecell{Precuneus\\Anterior} & Putamen & Thalamus & & \\
\midrule
\multirow{4}{*}{\makecell{Intra-scan\\Ridge}} & Canonical Fourier & 0.400$\pm$0.287 & 0.320$\pm$0.292 & 0.494$\pm$0.257 & 0.501$\pm$0.274 & 0.347$\pm$0.306 & 0.410$\pm$0.363 & 0.550$\pm$0.242 & 0.432 \\
 & Uniform Fourier & \secondcell{0.412$\pm$0.236} & \secondcell{0.370$\pm$0.268} & 0.510$\pm$0.277 & \secondcell{0.522$\pm$0.247} & \secondcell{0.393$\pm$0.271} & \secondcell{0.423$\pm$0.316} & \secondcell{0.561$\pm$0.248} & \secondcell{0.456} \\
 & Whole-history spectrum & 0.170$\pm$0.266 & 0.285$\pm$0.256 & \secondcell{0.524$\pm$0.244} & 0.447$\pm$0.270 & 0.274$\pm$0.334 & 0.327$\pm$0.363 & 0.513$\pm$0.256 & 0.363 \\
 & MEL--Morlet & \bestcell{0.437$\pm$0.267} & \bestcell{0.388$\pm$0.234} & \bestcell{0.604$\pm$0.211} & \bestcell{0.555$\pm$0.325} & \bestcell{0.472$\pm$0.243} & \bestcell{0.483$\pm$0.312} & \bestcell{0.615$\pm$0.199} & \bestcell{0.508} \\
\midrule
\multirow{4}{*}{\makecell{Intra-scan\\MLP}} & Canonical Fourier & \secondcell{0.360$\pm$0.185} & 0.236$\pm$0.211 & \secondcell{0.426$\pm$0.227} & \secondcell{0.418$\pm$0.273} & 0.302$\pm$0.230 & \secondcell{0.412$\pm$0.241} & 0.466$\pm$0.236 & \secondcell{0.374} \\
 & Uniform Fourier & 0.354$\pm$0.171 & \secondcell{0.240$\pm$0.229} & 0.410$\pm$0.226 & 0.405$\pm$0.241 & \secondcell{0.305$\pm$0.203} & 0.389$\pm$0.221 & \secondcell{0.477$\pm$0.201} & 0.368 \\
 & Whole-history spectrum & 0.218$\pm$0.233 & 0.190$\pm$0.214 & 0.388$\pm$0.205 & 0.368$\pm$0.236 & 0.227$\pm$0.286 & 0.329$\pm$0.253 & 0.393$\pm$0.243 & 0.302 \\
 & MEL--Morlet & \bestcell{0.385$\pm$0.233} & \bestcell{0.267$\pm$0.219} & \bestcell{0.513$\pm$0.209} & \bestcell{0.467$\pm$0.271} & \bestcell{0.394$\pm$0.199} & \bestcell{0.476$\pm$0.220} & \bestcell{0.528$\pm$0.245} & \bestcell{0.433} \\
\midrule
\multirow{4}{*}{\makecell{Intra-scan\\Ensemble}} & Canonical Fourier & 0.414$\pm$0.196 & \secondcell{0.292$\pm$0.251} & \secondcell{0.486$\pm$0.241} & \secondcell{0.478$\pm$0.300} & 0.346$\pm$0.259 & \secondcell{0.472$\pm$0.261} & 0.529$\pm$0.239 & \secondcell{0.431} \\
 & Uniform Fourier & \secondcell{0.417$\pm$0.193} & 0.290$\pm$0.269 & 0.473$\pm$0.248 & 0.474$\pm$0.268 & \secondcell{0.358$\pm$0.218} & 0.446$\pm$0.230 & \secondcell{0.548$\pm$0.212} & 0.429 \\
 & Whole-history spectrum & 0.263$\pm$0.268 & 0.239$\pm$0.256 & 0.466$\pm$0.227 & 0.435$\pm$0.276 & 0.259$\pm$0.353 & 0.373$\pm$0.289 & 0.459$\pm$0.270 & 0.356 \\
 & MEL--Morlet & \bestcell{0.434$\pm$0.255} & \bestcell{0.320$\pm$0.246} & \bestcell{0.578$\pm$0.211} & \bestcell{0.528$\pm$0.298} & \bestcell{0.449$\pm$0.203} & \bestcell{0.538$\pm$0.225} & \bestcell{0.590$\pm$0.248} & \bestcell{0.491} \\
\midrule
\multirow{3}{*}{\makecell{Inter-subject\\Ridge}} & Time-mean Fourier & 0.090$\pm$0.316 & \secondcell{0.115$\pm$0.222} & 0.293$\pm$0.255 & 0.235$\pm$0.261 & 0.176$\pm$0.302 & 0.210$\pm$0.339 & 0.245$\pm$0.312 & 0.195 \\
 & Fixed-weight Fourier & \secondcell{0.124$\pm$0.275} & 0.097$\pm$0.262 & \secondcell{0.337$\pm$0.228} & \secondcell{0.258$\pm$0.307} & \secondcell{0.196$\pm$0.305} & \secondcell{0.250$\pm$0.284} & \secondcell{0.271$\pm$0.308} & \secondcell{0.219} \\
 & Lagged Fourier & \bestcell{0.354$\pm$0.307} & \bestcell{0.257$\pm$0.266} & \bestcell{0.410$\pm$0.248} & \bestcell{0.476$\pm$0.240} & \bestcell{0.320$\pm$0.264} & \bestcell{0.366$\pm$0.278} & \bestcell{0.456$\pm$0.261} & \bestcell{0.377} \\
\bottomrule
\end{tabular}}
\caption{Regional prediction under matched readouts. Intra-scan entries are mean$\pm$standard deviation across 29 scans. MLP scores average five seeds within scan; ensemble scores average predictions before scoring. All intra-scan inputs contain 1040 features, with common settings within each readout family and 33 Ridge penalties. Inter-subject rows exclude every scan of the evaluated participant, using training-participant calibration and 31 target-specific penalties. Red and blue denote the best and second-best values within each block, not statistical significance.}
\label{tab:main_results}
\end{table*}
% END EMBEDDED table_main_results.tex

\subsubsection{External Oddball cohort}
External evaluation uses the auditory and visual Oddball EEG--fMRI dataset: 102 unique runs from 17 participants, 34 EEG channels, 170 target times per run, and 64 regional targets \cite{walz2018oddball}. Five participant-grouped folds assess unseen participants, while six run-key holdouts assess unseen runs from participants represented in training. Feature standardization is fitted within each training fold, and Ridge uses fixed penalty 10. Canonical Fourier models are refitted on Oddball; this analysis tests the representation recipe under an external protocol, not direct transfer of a fitted resting-state model. External Morlet estimation is not evaluated.

\subsection{Metrics and statistical analysis}
Pearson correlation is computed for each target and averaged over targets within a scan. Descriptive tables report the mean and standard deviation across scans. Paired inference first averages repeated scans within participant, yielding 22 equally weighted participant scores. Unless otherwise specified, 95\% percentile bootstrap confidence intervals (CIs) use 20,000 participant resamples with seed 20261011. Target-wise Morlet--canonical differences use the same pairing and resampling; these regional CIs are descriptive and unadjusted for target-wise multiplicity.

Exact two-sided sign tests assess the directional consistency of participant-paired differences. Holm correction is applied separately within four comparison families: 11 matched-readout contrasts against Morlet--Ridge, 14 estimator-and-aggregation contrasts, seven canonical temporal contrasts, and two participant-exclusion contrasts. Bootstrap CIs are not multiplicity-adjusted. Because test partitions were inspected during development, all comparisons are exploratory. Neither repeated scans, output targets, nor MLP seeds are treated as additional independent participants.

The paired-mean CI and sign test address different estimands: the former concerns the magnitude of the average difference, while the latter concerns its direction across participants. Large gains in a subset of participants can therefore produce a positive mean CI without a significant adjusted sign test. Correlation is the common prediction endpoint; canonical component ablations additionally report MSE on the normalized target scale. Without an independent reliability or noise-ceiling analysis, neither endpoint quantifies the fraction of recoverable BOLD variation. Aggregation and multiplicity procedures are detailed in \ref{app:statistics}.

\section{Results}
\subsection{Regional prediction and participant-level differences}
Under the matched Ridge protocol, Morlet features achieved mean scan-wise $R=0.508$, compared with 0.432 for canonical Fourier features, 0.456 for uniform Fourier features, and 0.363 for the whole-history spectrum (Table~\ref{tab:main_results}). After averaging repeated scans within participant, the Morlet mean was 0.506 (95\% CI, 0.444--0.565). The paired improvement over canonical Fourier features was 0.069 (95\% CI, 0.037--0.101), with gains in 19 of 22 participants and Holm-adjusted sign-test $p=0.006$. The estimator contrast was therefore evident with a linear readout, without adding a nonlinear mapping or prediction ensemble.

\begin{figure*}[pos=t]
\centering
\includegraphics[width=\textwidth]{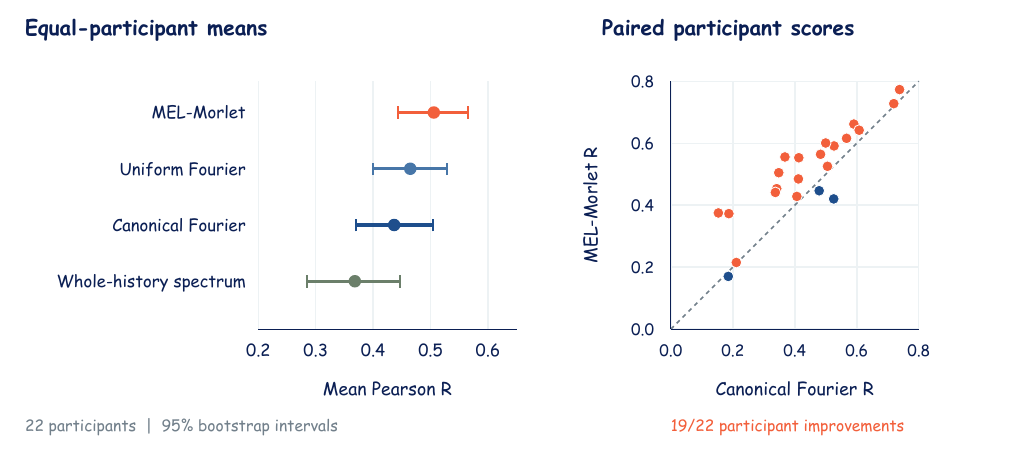}
\caption{Participant-level Ridge prediction after averaging repeated scans ($n=22$). Left: means and 95\% bootstrap CIs. Right: paired canonical Fourier and Morlet scores; the diagonal indicates equal performance.}
\label{fig:paired}
\end{figure*}

Figure~\ref{fig:paired} shows the distribution of participant-level scores and the paired Morlet--canonical comparison. The predominance of points above the equality line supports an improvement shared across participants rather than one attributable solely to the participants with repeated scans.

The scan-wise difference of approximately 0.076 and participant-paired difference of 0.069 reflect different weighting conventions: the former weights each of 29 scans equally, while the latter weights each of 22 participants equally after combining repeated scans. They describe the same comparison and are not independent confirmations. Uncertainty for the principal contrast is based on the participant-paired analysis.

The separate estimator-and-aggregation analysis yielded Morlet and canonical scores of 0.507 and 0.431. Its Morlet--canonical contrast retained adjusted $p=0.009$ across 14 comparisons. Comparisons with whole-history spectral and frequency-pooled Morlet features also passed the adjusted sign tests. The uniform Fourier contrast was less consistent across participants: its paired-mean CI was positive, but its adjusted sign test was inconclusive (Supplementary Tables S1--S2). These outcomes provide a sensitivity analysis under a different penalty grid, not an independent replication.

Prediction and estimator gains varied across targets. Morlet--Ridge achieved scan-wise correlations of 0.437 in the cuneus and 0.388 in Heschl's gyrus, compared with 0.604 in the anterior middle frontal gyrus and 0.615 for the global signal. Participant-paired gains were largest in the putamen (0.111) and middle frontal target (0.105), whereas the cuneus gain was 0.027 (Figure~\ref{fig:readout_targets}, right). Thus, a higher aggregate score did not imply a uniform regional improvement.

The regional pattern also differed in directional consistency. Putamen and middle frontal scores improved in 21 and 19 participants, respectively; cuneus and Heschl's gyrus scores each improved in 12. The anterior precuneus and thalamus CIs included zero despite positive mean differences. All seven paired estimates are reported in \ref{app:regional}. These descriptive results identify where the estimator contrast is more evident, but do not establish a region-specific physiological explanation.

\subsection{Representation effects across matched readouts}
\begin{figure*}[pos=t]
\centering
\includegraphics[width=\textwidth]{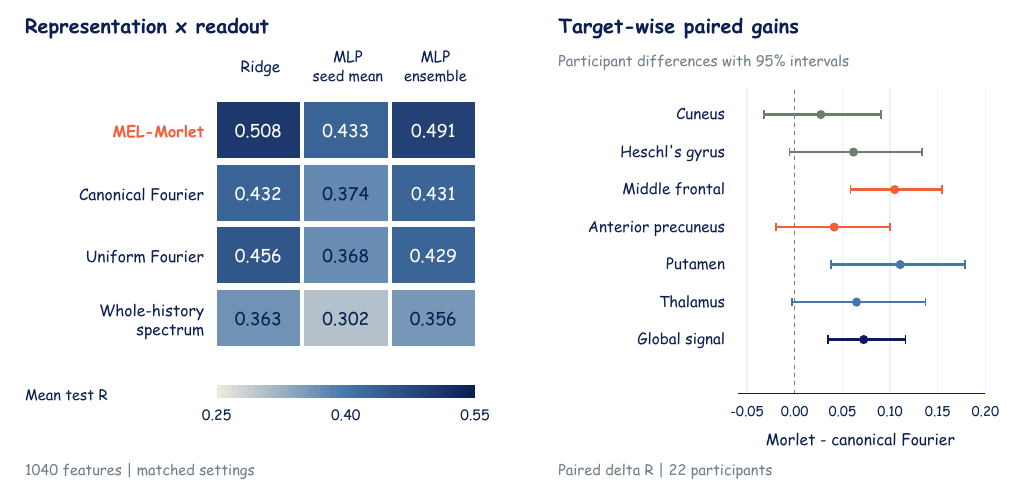}
\caption{Readout and regional comparisons. Left: mean correlation across 29 scans for each representation--readout combination. Right: participant-paired Morlet-minus-canonical Ridge differences ($n=22$) with 95\% bootstrap CIs, unadjusted for target-wise multiplicity.}
\label{fig:readout_targets}
\end{figure*}

Morlet features had the highest mean score within each matched readout family (Figure~\ref{fig:readout_targets}). Their MLP achieved seed-averaged $R=0.433$, while averaging the five prediction vectors yielded 0.491. Canonical and uniform Fourier ensembles reached 0.431 and 0.429, respectively. Ensembling improved the MLP endpoint, but was not necessary for the Morlet advantage over canonical inputs, which was already present under Ridge.

The participant-paired difference between Morlet--Ridge and the Morlet MLP ensemble was 0.019 (95\% CI, $-0.022$ to 0.069; adjusted $p=0.832$). This comparison does not establish superiority or equivalence between the two readouts. It shows that the estimator gain is obtainable with a low-capacity linear predictor; benefits from other nonlinear architectures remain an open question.

The strength of evidence depended on the comparator. Relative to uniform Fourier--Ridge, the participant-paired Morlet advantage was 0.041 (95\% CI, 0.005--0.076), but only 14 of 22 participants improved and the adjusted sign test was inconclusive ($p=0.573$). Relative to the single Morlet MLP, the advantage was 0.077, positive in 19 participants, with adjusted $p=0.006$. These contrasts distinguish the magnitude of a mean gain from its consistency across participants (\ref{app:statistics}). The strongest evidence concerns Morlet versus canonical Fourier under Ridge, rather than superiority over every alternative.

\subsection{Contribution of retained coordinates}
\subsubsection{Temporal, spectral, and channel aggregation}
Canonical component ablations quantify the effects of reducing spectral and temporal distinctions using both correlation and normalized-target MSE (Table~\ref{tab:component_ablation}). Five-band lagged Fourier features achieved $R=0.432$, compared with 0.391 for theta--alpha--beta features, 0.318 for fixed temporal weighting, and 0.261 for time averaging. Full five-band features also had the lowest MSE for six of seven targets; fixed weighting had the lowest MSE in Heschl's gyrus. The MSE results complement correlation by evaluating errors on the normalized BOLD amplitude scale.

% BEGIN EMBEDDED table_component_ablation.tex
\begin{table*}[pos=t]
\centering
\scriptsize
\setlength{\tabcolsep}{3.0pt}
\renewcommand{\arraystretch}{1.10}
\resizebox{\textwidth}{!}{
\begin{tabular}{c|cc|cc|cc|cc|cc|cc|cc|c}
\toprule
\multirow{3}{*}{\textbf{Model}} & \multicolumn{4}{c|}{\textbf{Primary Sensory}} & \multicolumn{4}{c|}{\textbf{High-level Cognitive}} & \multicolumn{4}{c|}{\textbf{Subcortical}} & \multicolumn{2}{c|}{\textbf{Global}} & \multirow{3}{*}{\textbf{Avg. $R\uparrow$}} \\
\cmidrule(lr){2-5}\cmidrule(lr){6-9}\cmidrule(lr){10-13}\cmidrule(lr){14-15}
& \multicolumn{2}{c|}{Cuneus} & \multicolumn{2}{c|}{Heschl's Gyrus} & \multicolumn{2}{c|}{Middle Frontal} & \multicolumn{2}{c|}{Precuneus Anterior} & \multicolumn{2}{c|}{Putamen} & \multicolumn{2}{c|}{Thalamus} & \multicolumn{2}{c|}{Global Signal} & \\
& MSE$\downarrow$ & $R\uparrow$ & MSE$\downarrow$ & $R\uparrow$ & MSE$\downarrow$ & $R\uparrow$ & MSE$\downarrow$ & $R\uparrow$ & MSE$\downarrow$ & $R\uparrow$ & MSE$\downarrow$ & $R\uparrow$ & MSE$\downarrow$ & $R\uparrow$ & \\
\midrule
$T_2$ & 0.431 & 0.164 & 0.433 & 0.133 & 0.341 & 0.347 & 0.344 & 0.360 & 0.393 & 0.198 & 0.350 & 0.262 & 0.362 & 0.366 & 0.261 \\
$T_1$ & 0.405 & 0.182 & \textbf{0.290} & 0.238 & 0.320 & 0.420 & 0.316 & 0.404 & 0.334 & 0.234 & 0.323 & 0.316 & 0.274 & 0.431 & 0.318 \\
$M+B_3$ & 0.324 & 0.370 & 0.361 & 0.253 & 0.361 & 0.455 & 0.334 & 0.453 & 0.371 & 0.339 & 0.344 & 0.383 & 0.325 & 0.485 & 0.391 \\
$M+B_5$ & \textbf{0.273} & \textbf{0.400} & 0.301 & \textbf{0.320} & \textbf{0.299} & \textbf{0.494} & \textbf{0.262} & \textbf{0.501} & \textbf{0.306} & \textbf{0.347} & \textbf{0.301} & \textbf{0.410} & \textbf{0.251} & \textbf{0.550} & \textbf{0.432} \\
\bottomrule
\end{tabular}}
\caption{Canonical Fourier component ablations, averaged across 29 scans under validation-selected Ridge. $T_2$: lag averaging; $T_1$: Gaussian temporal weighting centered at 7 seconds with width 2.5 seconds; $M+B_3$: theta, alpha, and beta bands; $M+B_5$: all five bands. MSE uses normalized BOLD targets. Bold denotes the lowest MSE or highest correlation per column.}
\label{tab:component_ablation}
\end{table*}
% END EMBEDDED table_component_ablation.tex

In the separate Morlet analysis, lag pooling reduced $R$ from 0.507 to 0.339, and frequency pooling reduced it to 0.426. Channel pooling retained 0.481, with an adjusted sign test that did not establish a consistent deficit. The canonical Fourier temporal analysis instead favored channel pooling, with 0.449 versus 0.431 for full coordinates (Supplementary Table S5). The benefit of channel-specific features was therefore contingent on the estimator and fitting procedure, whereas temporal and frequency distinctions were more consistently useful in these comparisons.

Compact representations also retained substantial predictive information. PCA--Ridge with 64 components achieved 0.497, and ROI-wise prediction fusion achieved 0.506. Axis marginals had higher mean validation correlation than full Morlet coordinates but lower test correlation. Full Morlet should consequently be understood as the subject of the controlled estimator comparison, not as the globally validation-optimal representation among all evaluated variants.

Taken together, the reductions favor a temporally distributed spectral description while revealing redundancy in its full coordinate array. A compact projection can retain much of the prediction signal, and retaining channel identity is not uniformly beneficial. Because pooling changes dimensionality and regularization geometry as well as information content, the observed losses cannot be decomposed into additive contributions of lag, channel, and frequency.

\subsubsection{Coordinate consistency and feature--target correspondence}
Shared permutations preserved the aligned Morlet score of approximately 0.507 after refitting, with changes below $5\times10^{-6}$ (Table~\ref{tab:controls}). By contrast, independent per-example channel, frequency, and lag shuffles reduced scores to 0.476, 0.410, and 0.330. Shuffling all coordinates yielded 0.156, while circular feature--target shifts and shuffled training labels yielded 0.064 and 0.009, respectively.

% BEGIN EMBEDDED table_controls.tex
\begin{table}[pos=tb]
\centering
\scriptsize
\setlength{\tabcolsep}{2.5pt}
\renewcommand{\arraystretch}{1.08}
\begin{tabular}{llrr}
\toprule
\textbf{Control} & \textbf{Mapping} & $R\uparrow$ & $\Delta R$ \\
\midrule
Aligned & None & 0.507 & +0.000 \\
Fixed channel & Reindexed & 0.507 & +0.000 \\
Fixed lag & Reindexed & 0.507 & +0.000 \\
Fixed flat & Reindexed & 0.507 & +0.000 \\
Fixed band & Reindexed & 0.507 & +0.000 \\
Sample channel & Channel identity & 0.476 & -0.031 \\
Sample band & Frequency identity & 0.410 & -0.097 \\
Sample lag & Lag identity & 0.330 & -0.178 \\
Sample all & All identities & 0.156 & -0.351 \\
Circular shift & Target pairing & 0.064 & -0.443 \\
Label shuffle & Training pairing & 0.009 & -0.498 \\
\bottomrule
\end{tabular}
\caption{Refitted coordinate controls in the 31-penalty Morlet analysis. Shared transformations reindex all splits consistently; sample-wise controls independently disrupt each example. $\Delta R$ is relative to aligned features. Shared-permutation changes are below $5\times10^{-6}$ and vanish at the displayed precision.}
\label{tab:controls}
\end{table}
% END EMBEDDED table_controls.tex

The shared-permutation results agree with the algebraic invariance of Ridge: changing a common feature order changes neither the fitted function nor its penalty after corresponding reindexing. The tensor therefore provides an identifiable organization of measurements rather than a geometric constraint on the flat predictor.

Independent per-example permutations instead remove a common mapping between measurement identity and coefficient, while feature--target shifts disrupt history--response pairing. Their larger effects support the importance of consistent correspondence across observations. They do not show that Ridge exploits adjacency in the stored tensor or that a disrupted coordinate uniquely represents a physiological process. The invariance argument and intervention scope are developed in \ref{app:coordinates}.

\subsubsection{Temporal resolution and cached support}
\begin{figure*}[pos=t]
\centering
\includegraphics[width=\textwidth]{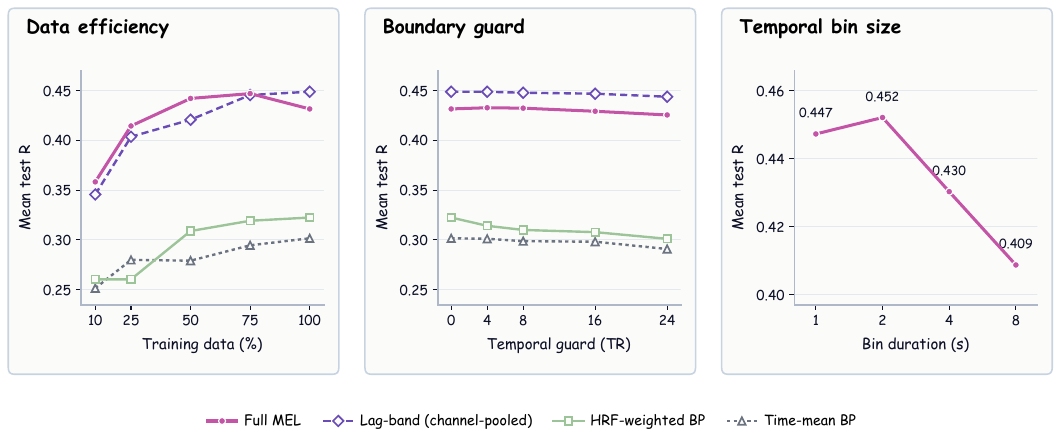}
\caption{Canonical Fourier sensitivity to training fraction (left), additional split guards (middle), and bin duration (right). The first two panels use target-specific penalties; the last uses a shared penalty. Training subsets are independently sampled across representations. No uncertainty intervals are shown.}
\label{fig:sensitivity}
\end{figure*}

Canonical Fourier prediction varied modestly across the additional cache guards, with scores of approximately 0.425--0.433 (Figure~\ref{fig:sensitivity}). Under historical shared-penalty selection, two-second bins achieved 0.452, compared with 0.430 and 0.409 for four- and eight-second bins and 0.447 for one-second bins. This pattern is consistent with useful within-history temporal resolution. The shared-penalty scores are not directly interchangeable with the target-specific result of 0.432 in Table~\ref{tab:main_results}.

Training-fraction curves characterize sensitivity within each representation because their subsets were sampled independently. They cannot establish paired differences in sample efficiency. Likewise, larger cache guards probe sensitivity to partition spacing within the inherited pipeline; they do not remove dependencies introduced by whole-scan BOLD preprocessing.

\subsection{Participant-separated and external prediction}
Under participant exclusion, canonical lagged Fourier regression achieved scan-wise $R=0.377$, compared with 0.219 for fixed weighting and 0.195 for time averaging (Table~\ref{tab:main_results}). Its participant mean was 0.377 (95\% CI, 0.299--0.454). The paired advantage over time averaging was 0.167 (95\% CI, 0.099--0.235; adjusted $p=0.009$), whereas the fixed-weighting contrast did not pass the adjusted sign test. Retaining temporal distinctions was therefore useful without fitting to the evaluated participant, although absolute prediction was lower than in the intra-scan canonical analysis.

External Oddball prediction was weak under both holdout protocols (Table~\ref{tab:external}). Canonical lagged regression reached 0.026 for unseen participants and 0.091 for held-out runs from represented participants; time averaging reached 0.018 and 0.088. These low absolute correlations and small descriptive differences do not establish effective external participant prediction. Because only canonical models were refitted in this analysis, the result constrains the demonstrated portability of that representation recipe without directly evaluating external Morlet performance.

These settings answer distinct generalization questions. Intra-scan evaluation uses an individual's earlier observations to predict later observations retrospectively. Participant exclusion evaluates a pooled readout without that individual's fitting data. Oddball evaluates a refitted representation recipe under different acquisition, task, and target definitions. The scores should not be treated as a single nested transfer experiment: the available training data change, and the seven-target resting-state endpoint differs from the 64-target Oddball endpoint.

% BEGIN EMBEDDED table_external.tex
\begin{table}[pos=tb]
\centering
\footnotesize
\setlength{\tabcolsep}{3.0pt}
\renewcommand{\arraystretch}{1.12}
\begin{tabular}{lrr}
\toprule
\textbf{Representation} & \makecell{Unseen\\participant} & \makecell{Held-out\\run} \\
\midrule
Time-mean Fourier & 0.018 & 0.088 \\
Fixed-weight Fourier & 0.015 & 0.072 \\
Lagged Fourier & 0.026 & 0.091 \\
\bottomrule
\end{tabular}
\caption{External Oddball prediction across 102 runs from 17 participants. Correlations are averaged over 64 targets within run, then over runs. Models use fold-specific training calibration and fixed Ridge penalty 10. Temporal weighting uses a gamma kernel (shape 6, scale 1), rather than the resting-state Gaussian kernel.}
\label{tab:external}
\end{table}
% END EMBEDDED table_external.tex

\section{Discussion}
The central finding is that a change in spectral measurement improved regional BOLD prediction under a fixed linear readout. The participant-paired Morlet--canonical contrast was accompanied by consistent gains across most participants, while matched nonlinear comparisons showed that the advantage did not depend on adding an MLP or prediction ensemble. The interpretation is therefore a contribution from measurement design within the evaluated pipeline, rather than evidence for a new learning architecture.

The complementary analyses clarify what this contribution entails. Aggregation indicates that temporal and spectral distinctions carry useful prediction information, while refitted controls distinguish stable feature identities from arbitrary coordinate order. Competitive compact representations show that the full tensor is not uniquely necessary, and weak external prediction limits the portability demonstrated by the current experiments. Together, these findings support an explicit interface for controlled modeling, with conclusions restricted to the measurements and evaluation protocols actually tested.

\subsection{Spectral measurements and the value of coordinate retention}
Morlet features achieved the highest scan mean within each matched readout family (Figure~\ref{fig:readout_targets}). Because architecture and training settings were fixed within a family, the observed differences are attributable to the representation choice within that specification, rather than to a larger decoder or a different ensemble size. Morlet bandwidth, taper, and frequency-dependent temporal support jointly change the energy summaries available to the predictor. They provide plausible explanations for the gain, but the present comparison does not identify which property is responsible. Resolving that question requires controlled variation of the individual estimator components.

The greater losses under temporal and frequency pooling are compatible with delayed, rhythm-dependent EEG--BOLD associations \cite{mantini2007signatures,scheeringa2011neuronal}. This compatibility is not a mechanistic demonstration: pooling also changes feature dimension and the regularized fitting problem. Likewise, permutation invariance shows that useful coordinate definitions need not impose geometric structure on a flat readout. The evidence supports retaining identifiable temporal--spectral distinctions while leaving their physiological origins unresolved.

The 1040-coordinate tensor is best viewed as an inspectable reference, not a claim about the minimum sufficient representation. Competitive channel-pooled and PCA results indicate redundancy that may be exploited after measurement. A compact representation can therefore preserve useful information without retaining every original coordinate. Conversely, a pooling loss does not establish physiological interactions among the retained axes. A factorial comparison of estimator support, aggregation, and regularization would be needed to separate these contributions more precisely.

\subsection{Implications for biomedical signal modeling}
An explicit spectral interface provides a transparent reference before introducing a larger prediction model. Its operations are specified numerically, and each fitted input coefficient can be traced to a measurement definition. This complements learned biosignal representations, which can encode temporal and channel information through positional embeddings or structured attention \cite{yang2023biot}. The distinction is direct measurement-level interpretability of the input coordinates, not the presence of structure in MEL and its absence from learned encoders.

Separating alignment, spectral estimation, calibration, and readout also makes the source of a performance change easier to identify. Replacing an estimator changes the supplied measurements; replacing a readout changes the fitted mapping; ensembling changes the final prediction rule. Maintaining these distinctions is particularly useful in small biomedical cohorts, where broad architectural searches can consume limited validation information. The interface supports controlled benchmarking, while independent data are still required to establish generalizability.

The results motivate a benchmarking sequence in which spectral estimators and compact summaries are established before additional readout complexity is assessed under a common selection budget. The paired CI for Ridge versus the Morlet MLP ensemble includes zero, providing no basis for a general claim of linear superiority. Nor does this study establish architecture-level ranking against NeuroBOLT, which requires matched retraining \cite{li2024neurobolt}. Whole-volume image metrics from NT-ViT and E2fNet address different endpoints and cannot be ranked directly against seven-target time-course correlations \cite{lanzino2024ntvit,roos2025e2fnet}.

The regional breakdown is equally important for biomedical interpretation. Larger middle frontal and putamen gains coexist with less certain sensory-region improvements. Differences in target reliability, shared slow fluctuations, and region-specific associations are all compatible with this pattern. Without reliability measurements or controlled physiological analyses, the observed heterogeneity cannot distinguish these explanations. The present endpoints concern regional signal prediction; scalp interpolation and regression accuracy do not establish cortical sources, diagnostic utility, or treatment benefit.

\subsection{Participant variability and the external generalization boundary}
Canonical Fourier prediction under participant exclusion remained below its intra-scan level, indicating that within-scan utility did not translate fully to a pooled cross-participant model. Its advantage over time averaging nevertheless supports retaining a distributed history in that setting. External Oddball scores were substantially lower in absolute terms, even after refitting the canonical models. Because target construction and acquisition also differed, this is a limitation on demonstrated external utility, not an estimate of the performance loss caused solely by changing participants or of external Morlet generalization.

Channel configuration, synchronization, task-related dynamics, regional target construction, recording quality, and BOLD reliability could each affect external prediction. The low scores alone do not identify their relative contributions. A harmonized external analysis should evaluate both estimators under participant separation while measuring acquisition quality and target reliability. Until such evidence is available, the demonstrated scope remains retrospective regional modeling in the primary cohort, with limited evidence of portability under the external protocol.

\subsection{Limitations and future directions}
The primary cohort comprises 22 participants from a single source, and each intra-scan test partition contains only 48 temporally dependent observations. Participant-level resampling avoids treating repeated scans as independent participants but cannot overcome limited cohort size or short test duration. In addition, inspection of the test partitions during development makes the reported comparisons exploratory. Confirmation of the estimator gain requires an untouched participant-separated cohort and a frozen analysis recipe.

Response preprocessing imposes a distinct limitation. Whole-scan filtering and normalization introduce dependencies that are not removed by training-only feature calibration. Larger cache guards alter partition spacing, and participant exclusion separates people, but neither reconstructs the BOLD preprocessing within the intended prediction protocol. A prospective interpretation therefore requires rebuilding the raw-data pipeline with split-aware processing before repeating the matched comparisons.

The comparison also has architectural and measurement limits. Equal feature counts do not equate bandwidth, temporal support, or effective readout degrees of freedom. The matched MLP is a specified compact nonlinear reference rather than an exhaustive neural benchmark. Paired MSE is available for the canonical component ablations, but not as a complete amplitude-error comparison across Morlet, matched MLP, and external experiments. End-to-end runtimes, noise ceilings, and independent validation of coefficient maps are also unavailable. The current evidence consequently supports claims about correlation under the stated protocols, not established superiority in efficiency, amplitude fidelity, or physiological explanation.

The next priority is independent validation rather than further optimization on the existing test partitions. After freezing the estimator and selection procedure, evaluation on untouched participant-separated recordings with split-aware preprocessing can establish whether the principal contrast persists. Controlled kernel comparisons and matched larger readouts can then address measurement and architecture separately. Recording-quality diagnostics and reliability-aware errors would clarify the attainable signal-prediction accuracy; clinical endpoints would be required before considering a biomedical decision-making application.

\section{Conclusions}
MEL provides an explicit interface from pre-target EEG histories to regional BOLD prediction through identifiable spectral measurements and regularized readouts. In the studied resting-state cohort, Morlet estimation improved on canonical Fourier features by a participant-paired mean correlation difference of 0.069 under Ridge. Matched-readout comparisons locate this gain in the measurement choice without requiring additional nonlinear capacity, while aggregation and refitted controls delimit the interpretation of coordinate retention. The resulting interface is a transparent reference for cross-modal modeling, not an established general-purpose EEG-to-fMRI solution. Weak external prediction and retrospective preprocessing make independent, participant-separated validation with split-aware processing essential.

\section*{Data and analysis availability}
The resting-state data are distributed by the NeuroBOLT authors at \url{https://huggingface.co/datasets/NeurdyLab/NeuroBOLT}. The external data are associated with the Oddball record ds000116 \cite{walz2018oddball}. Access and redistribution are governed by the original data conditions; raw recordings are not redistributed here. The appendices specify the measurement interface, selection procedures, paired statistics, and display definitions. Complete experiment-family score tables and provenance are retained in the supplementary analysis files.

\appendix
\renewcommand{\thesection}{Appendix~\Alph{section}}
\renewcommand{\thesubsection}{\Alph{section}.\arabic{subsection}}
\numberwithin{equation}{section}
\numberwithin{table}{section}
\numberwithin{figure}{section}
\renewcommand{\theequation}{\Alph{section}.\arabic{equation}}
\renewcommand{\thetable}{\Alph{section}.\arabic{table}}
\renewcommand{\thefigure}{\Alph{section}.\arabic{figure}}

\section{Numerical specification of the measurement interface}
\label{app:implementation}
\subsection{History, segmentation, and Fourier coordinates}
Each resting-state input contains $26\times3200$ EEG samples. Zero-based lag index $l$ identifies the interval $[t_n-16+2l,t_n-16+2(l+1))$, whose signal for each channel is a 400-element vector. Segmentation produces an $8\times26\times400$ tensor, and spectral reduction yields $8\times26\times5$ measurements. The 1040-element vector is ordered by lag, channel, and then frequency. Channel indices follow the retained recording order and do not encode a learned spatial arrangement.

The canonical Fourier estimator multiplies each 400-sample segment by a discrete Hann window and computes the squared magnitudes of its real FFT. Frequency samples are spaced by 0.5 Hz. Band membership includes the lower edge and excludes the upper edge. Squared magnitudes are averaged over the included frequency samples before log compression. Uniform Fourier measurements retain the temporal intervals and channel order but use frequency edges equally spaced from 0.5 to 45 Hz.

For the whole-history comparator, a Hann window spans all 3200 samples, and the squared Fourier magnitudes are pooled over 40 equally spaced intervals covering 0.5--45 Hz. Flattening the $26\times40$ array yields 1040 features. This matches the feature count by exchanging temporal distinctions for finer frequency subdivisions. None of the Fourier implementations divides by window energy and sampling rate to produce a density-normalized power spectral density (PSD).

\subsection{Morlet kernels and history boundaries}
For center frequency $\nu_f$ and cycle count $q_f$, set $\sigma_f=q_f/(2\pi\nu_f)$ and $a_f=\max(8,\lceil3.5f_s\sigma_f\rceil)$. At integer sample offsets $k=-a_f,\ldots,a_f$, the unnormalized kernel and its normalized version are
\begin{equation}
\begin{aligned}
u_f[k]&=\exp(2\pi i\nu_f k/f_s)
         \exp\!\left[-\frac{(k/f_s)^2}{2\sigma_f^2}\right],\\
\psi_f[k]&=\frac{u_f[k]-\overline{u_f}}
 {\sqrt{\sum_j|u_f[j]-\overline{u_f}|^2+10^{-12}}}.
\end{aligned}
\label{eq:kernel_specification}
\end{equation}
Here $i^2=-1$, and $\overline{u_f}$ denotes the discrete kernel mean. The frequency--cycle pairs $(\nu_f,q_f)$ are $(2,3)$, $(6,4)$, $(10.5,5)$, $(21.5,7)$, and $(37.5,9)$. Kernels are stored as single-precision complex arrays. FFT convolution uses the smallest power-of-two length at least as large as $3200+2a_f$; cropping begins at offset $a_f$ and retains 3200 samples. Squared convolution magnitudes are grouped into eight consecutive 400-sample intervals, averaged within interval, and compressed by $\log(1+p)$.

Convolution assumes zero values outside the available history rather than using EEG beyond its boundaries. Centered kernels can use later samples relative to a location \emph{within} that history, so the operation is not a sample-by-sample causal filter. It nevertheless respects the target-relative input boundary because the entire history precedes $t_n$. Edge effects at the oldest and most recent samples are part of the evaluated estimator and have not been isolated in a separate boundary-handling experiment.

\subsection{Calibration and fitted parameter storage}
Training means and population standard deviations are estimated separately for each log-transformed coordinate. Standard deviations below $10^{-6}$ are replaced by one, and the resulting statistics are applied unchanged to validation and test arrays. Vectorization follows $j=(lC+c)F+f$. Inference applies the stored estimator settings, calibration statistics, and readout parameters without refitting normalization or accessing target values.

The seven-target Ridge readout stores $1040\times7=7280$ coefficients and seven intercepts. The MLP contains 2080 layer-normalization parameters, 16,656 first-layer parameters, and 119 output-layer parameters, totaling 18,855. The five-model ensemble stores five MLPs and averages their outputs. These readout counts exclude calibration statistics, cached histories, feature-extraction memory, and the workload of validation-grid fitting.

\section{Analysis families and validation selection}
\label{app:selection}
\subsection{Protocol ledger}
Table~\ref{tab:protocol_ledger} distinguishes the selection procedures used across analyses. The primary matched-readout comparison uses the 33-penalty configuration. The 31-penalty estimator analysis, 17-penalty canonical temporal analysis, historical shared-penalty diagnostics, and participant-exclusion analysis use different configurations. They reuse the resting-state cohort and provide complementary comparisons rather than independent replications.

\begin{table*}[pos=t]
\centering
\small
\setlength{\tabcolsep}{4pt}
\renewcommand{\arraystretch}{1.16}
\begin{tabular}{p{0.19\textwidth}p{0.26\textwidth}p{0.23\textwidth}p{0.24\textwidth}}
\toprule
\textbf{Analysis} & \textbf{Readout and selection} & \textbf{What is compared} & \textbf{Interpretation} \\
\midrule
Matched readouts & 33 Ridge penalties, $10^{-1}$--$10^7$, per target; fixed MLP with five seeds & Four equal-dimensional inputs within each readout family & Estimator and ensemble effects in the primary comparison \\
Estimator and aggregation & 31 Ridge penalties, $10^{-2}$--$10^8$, LSQR; validation-selected projections and fusion & Full Morlet, pooled coordinates, compact summaries, and spectral references & Retained information with changed feature dimensions \\
Canonical temporal controls & 17 Ridge penalties, $10^{-3}$--$10^5$, per target & Distributed, weighted, pooled, and smoothed Fourier summaries & Separate temporal-reference experiment \\
Historical bin sensitivity & 33 penalties, $10^{-1}$--$10^7$; one shared penalty across targets & One-, two-, four-, and eight-second bins & Bin-duration sensitivity, not the primary score \\
Frozen-readout sensitivity & Six penalties, $10^{-1}$--$10^4$; one shared validation-selected penalty & Forty canonical Fourier lag--band input ablations & Exploratory test-input sensitivity without refitting \\
Participant exclusion & 31 penalties, $10^{-2}$--$10^8$, per target; training-participant calibration & Lagged, time-mean, and fixed-weight Fourier features & Unseen-participant regression within the cohort \\
External Oddball & Fixed Ridge penalty 10; training-fold calibration & Three canonical temporal summaries & Refitted participant-grouped and run-key-held-out prediction \\
\bottomrule
\end{tabular}
\caption{Analysis configurations and interpretation. Ridge penalties and MLP checkpoints use validation responses, except that the external penalty is fixed. Distinct configurations can produce different scores on the reused cohort and do not constitute independent replications.}
\label{tab:protocol_ledger}
\end{table*}

For each Ridge target, validation correlation selects a fitted candidate, whose test prediction is then scored. MLP seeds are fixed at 1--5. Each seed retains the checkpoint with the highest mean validation correlation across seven targets. The seed-mean endpoint includes all five selected models; the ensemble endpoint averages their prediction vectors before computing correlation. Neither endpoint selects a seed using test performance.

Validation-based model selection is distinct from independence of the overall development process. The test partitions were inspected during earlier development, so the current analyses remain exploratory even though individual penalties and checkpoints are selected on validation responses. The ledger documents those within-experiment procedures without implying preregistration or an untouched test cohort.

\subsection{Aggregation, projections, and fusion}
In the Morlet analysis, channel pooling yields $LF=40$ features, frequency pooling yields $LC=208$, and lag pooling yields $CF=130$. Concatenated axis marginals contain $LF+CF+LC=378$ features. Channel means and standard deviations at each lag--frequency location yield $2LF=80$ features. Pooling is applied to calibrated coordinates, followed by training-only scaling of the reduced vectors before fitting.

PCA is fitted to training inputs; the selected setting retains 64 components without whitening. PLS uses training inputs and responses, with four selected latent components. Prediction fusion combines canonical Fourier, uniform Fourier, full Morlet, and channel-pooled Morlet outputs. Validation predictions select either a shared combination-weight vector or separate target-wise vectors. Complete descriptive and paired comparisons are reported in Supplementary Tables S1--S2.

Axis marginals have a mean validation correlation of approximately 0.558, compared with 0.543 for full Morlet coordinates, while their test correlations are approximately 0.460 and 0.507, respectively. ROI-wise fusion also has a higher mean validation score than the full representation. Full Morlet is therefore not the globally validation-optimal candidate among these variants. Its role in the main analysis is the controlled equal-dimensional estimator comparison.

\begin{figure*}[pos=tbp]
\centering
\includegraphics[width=\textwidth]{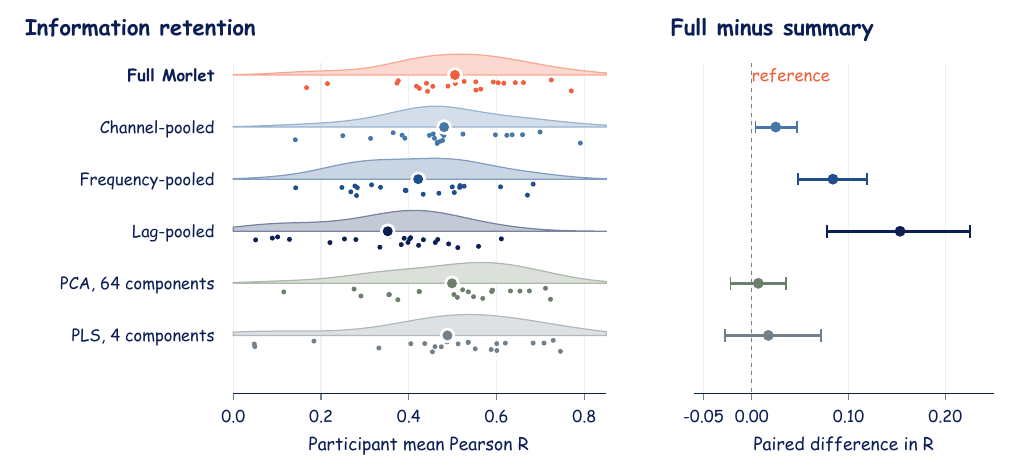}
\caption{Morlet aggregation and compression. Left: participant-score distributions, with small dots for participants and large dots for means ($n=22$). Right: full-minus-reduced paired differences with unadjusted 95\% bootstrap CIs. Repeated scans are averaged within participant.}
\label{fig:appendix_retention}
\end{figure*}

Figure~\ref{fig:appendix_retention} summarizes the separate 31-penalty analysis. Inference for full-versus-reduced representations is based on paired differences, not on whether their marginal score distributions overlap.

\subsection{Canonical temporal references}
The resting-state fixed-weight reference applies normalized Gaussian weights at lag midpoints, centered seven seconds before the target with width 2.5 seconds. Additional temporal references use the corresponding derivative and smoothing with weights $(0.25,0.5,0.25)$ and edge padding. These kernels define regression summaries; they are not participant-specific estimates of a physiological hemodynamic response function. Complete descriptive and paired results are retained in Supplementary Tables S5--S6.

Under historical shared-penalty selection, one-, two-, four-, and eight-second bins achieve 0.447, 0.452, 0.430, and 0.409, respectively. Extra cache guards of 0, 4, 8, 16, and 24 target times yield approximately 0.431, 0.433, 0.432, 0.429, and 0.425. These diagnostics describe sensitivity within the cached pipeline and do not replace rebuilding the upstream preprocessing.

\section{Participant aggregation and paired statistical evidence}
\label{app:statistics}
\subsection{Unit of analysis}
Let $R_{s,a,r}^{(m)}$ denote the test correlation for method $m$, participant $s$, scan $a$, and target $r$. With $q_s$ scans for participant $s$ and $K=7$ targets, the participant score and paired method difference are
\begin{equation}
\begin{aligned}
\overline R_s^{(m)}
 &=\frac{1}{q_s}\sum_{a=1}^{q_s}\frac{1}{K}
   \sum_{r=1}^{K} R_{s,a,r}^{(m)},\\
d_s^{(m,b)}&=\overline R_s^{(m)}-\overline R_s^{(b)}.
\end{aligned}
\label{eq:participant_aggregation}
\end{equation}
The overall participant score is the arithmetic mean of the 22 values $\overline R_s^{(m)}$; the paired mean difference is the arithmetic mean of $d_s^{(m,b)}$. Correlations are averaged on their original scale without Fisher transformation. Standard deviations across scans in the main table quantify dispersion and are not confidence intervals for the participant-level effect.

Percentile CIs use 20,000 resamples of 22 participants drawn with replacement using seed 20261011. Each resample retains the paired method difference for a participant rather than sampling the two methods independently. This preserves matching and prevents repeated scans from receiving additional participant-level weight. It does not account for all uncertainty associated with estimating a correlation from a short, temporally dependent test interval.

\subsection{Consistency, multiplicity, and selected contrasts}
The exact two-sided sign test counts positive and negative $d_s$, omits exact zeros, and evaluates the binomial tail under equal sign probabilities. Holm adjustment is applied within each complete family: 11 matched-readout contrasts, 14 estimator-and-aggregation contrasts, seven canonical temporal contrasts, and two participant-exclusion contrasts. The selected comparisons in Table~\ref{tab:paired_evidence} retain adjustment over their complete families rather than over only the displayed subset.

\begin{table*}[pos=t]
\centering
\small
\setlength{\tabcolsep}{5pt}
\renewcommand{\arraystretch}{1.16}
\begin{tabular}{p{0.44\textwidth}rrrr}
\toprule
\textbf{Reference minus comparator} & \textbf{Mean $\Delta R$} & \textbf{95\% interval} & \textbf{Gains/22} & \textbf{Adjusted $p$} \\
\midrule
Morlet--Ridge minus canonical--Ridge & +0.069 & [+0.037, +0.101] & 19/22 & 0.0060 \\
Morlet--Ridge minus uniform--Ridge & +0.041 & [+0.005, +0.076] & 14/22 & 0.5726 \\
Morlet--Ridge minus whole-history--Ridge & +0.138 & [+0.088, +0.196] & 20/22 & 0.0011 \\
Morlet--Ridge minus Morlet--MLP & +0.077 & [+0.041, +0.125] & 19/22 & 0.0060 \\
Morlet--Ridge minus Morlet--MLP ensemble & +0.019 & [-0.022, +0.069] & 12/22 & 0.8318 \\
\midrule
Full Morlet minus lag-pooled Morlet & +0.153 & [+0.078, +0.225] & 18/22 & 0.0391 \\
Full Morlet minus frequency-pooled Morlet & +0.084 & [+0.048, +0.119] & 20/22 & 0.0015 \\
Full Morlet minus channel-pooled Morlet & +0.025 & [+0.004, +0.047] & 14/22 & 1.0000 \\
\midrule
Participant-excluded lagged minus time-mean Fourier & +0.167 & [+0.099, +0.235] & 18/22 & 0.0087 \\
Participant-excluded lagged minus fixed-weight Fourier & +0.136 & [+0.068, +0.204] & 15/22 & 0.1338 \\
\bottomrule
\end{tabular}
\caption{Selected participant-paired contrasts. Blocks correspond to the 33-penalty matched-readout, 31-penalty estimator-and-aggregation, and participant-exclusion families. Positive differences favor the reference; gains count participants with positive differences. CIs are unadjusted, while sign-test $p$-values retain Holm correction over the complete family.}
\label{tab:paired_evidence}
\end{table*}

A positive paired-mean CI can coexist with an inconclusive adjusted sign test because the procedures address different properties. The CI incorporates the magnitudes of participant differences; the sign test uses their directions and is additionally subject to family-wise correction. This distinction explains the uniform Fourier and channel-pooled Morlet comparisons. All reported CIs remain conditional on the available cohort and exploratory development procedure.

\section{Complete target-wise estimator differences}
\label{app:regional}
For each target in the 33-penalty matched experiment, correlations are averaged over repeated scans within participant before forming Morlet-minus-canonical Ridge differences. The same 20,000 participant resamples yield percentile CIs. Table~\ref{tab:regional_evidence} provides the numerical counterpart of the regional-gain figure.

\begin{figure*}[pos=tbp]
\centering
\includegraphics[width=\textwidth]{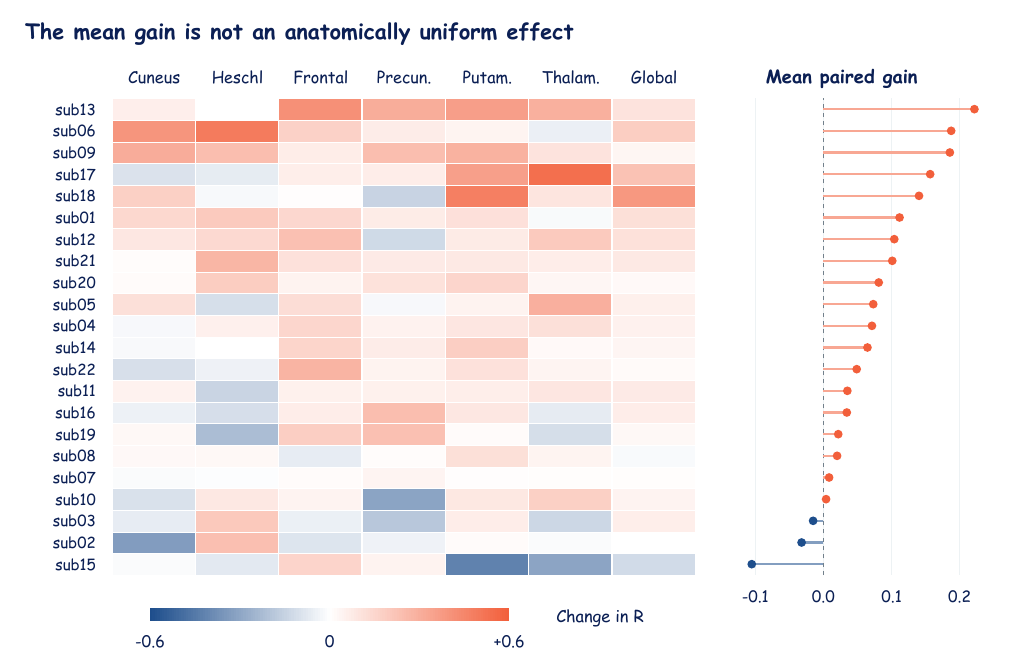}
\caption{Participant-by-target Morlet-minus-canonical Ridge correlation differences after averaging repeated scans. The heatmap uses a symmetric scale; the side plot shows seven-target mean differences and defines the row order. All 22 participants are included.}
\label{fig:appendix_target_atlas}
\end{figure*}

Figure~\ref{fig:appendix_target_atlas} displays the participant-by-target differences, including target-specific losses that are obscured by aggregate gains. Rows are ordered by the observed mean difference for display; this ordering is not an independent participant characteristic or clinical association.

\begin{table}[pos=tb]
\centering
\small
\setlength{\tabcolsep}{3pt}
\renewcommand{\arraystretch}{1.16}
\begin{tabular}{lrrr}
\toprule
\textbf{Target} & \textbf{$\Delta R$} & \textbf{95\% interval} & \textbf{Gains/22} \\
\midrule
Cuneus & +0.027 & [-0.033, +0.090] & 12/22 \\
Heschl's gyrus & +0.062 & [-0.005, +0.133] & 12/22 \\
Middle frontal & +0.105 & [+0.059, +0.155] & 19/22 \\
Anterior precuneus & +0.041 & [-0.019, +0.100] & 16/22 \\
Putamen & +0.111 & [+0.038, +0.179] & 21/22 \\
Thalamus & +0.065 & [-0.003, +0.137] & 15/22 \\
Global signal & +0.072 & [+0.035, +0.116] & 20/22 \\
\bottomrule
\end{tabular}
\caption{Target-wise Morlet-minus-canonical Ridge differences across 22 participants. Gains count positive participant differences. The 95\% bootstrap CIs are descriptive and unadjusted for target-wise multiplicity.}
\label{tab:regional_evidence}
\end{table}

The arithmetic mean of the seven unrounded target differences equals the overall paired difference of 0.069, linking the aggregate endpoint to the complete target breakdown. The larger putamen and middle frontal estimates and less certain sensory-region gains may reflect differences in target reliability, shared dynamics, or regional associations. Without corresponding reliability or physiological analyses, the intervals and anatomical grouping cannot distinguish these explanations.

\begin{figure*}[pos=tbp]
\centering
\includegraphics[width=\textwidth]{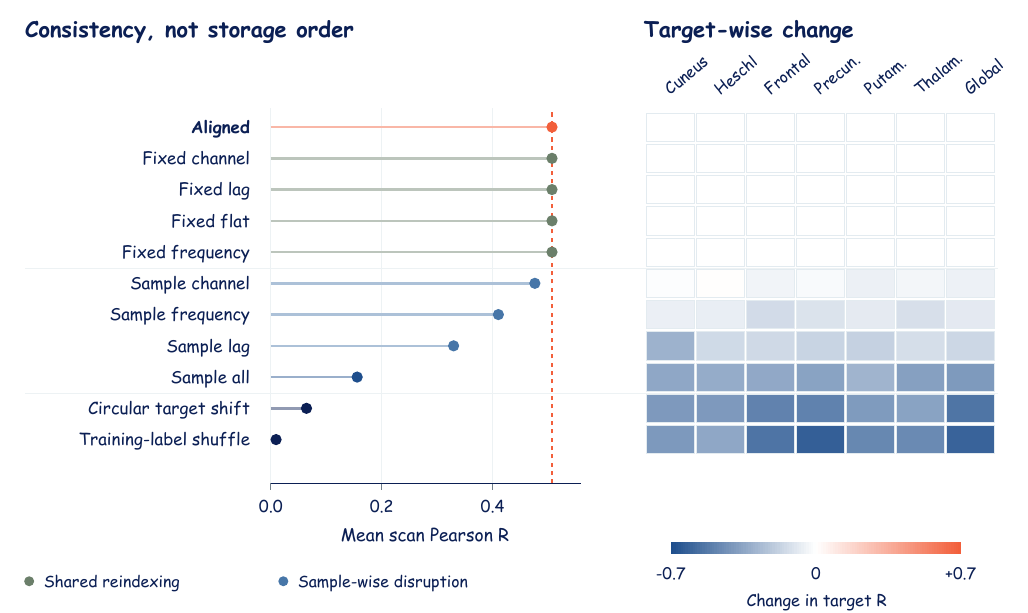}
\caption{Refitted Morlet correspondence controls. Left: mean correlation across 29 scans and seven targets, with the aligned baseline dashed. Right: target-wise changes on a shared symmetric scale; near-white cells indicate negligible change. Confidence intervals are unavailable.}
\label{fig:appendix_coordinate_controls}
\end{figure*}
\begin{figure*}[pos=p]
\centering
\includegraphics[width=\textwidth,height=0.70\textheight,keepaspectratio]{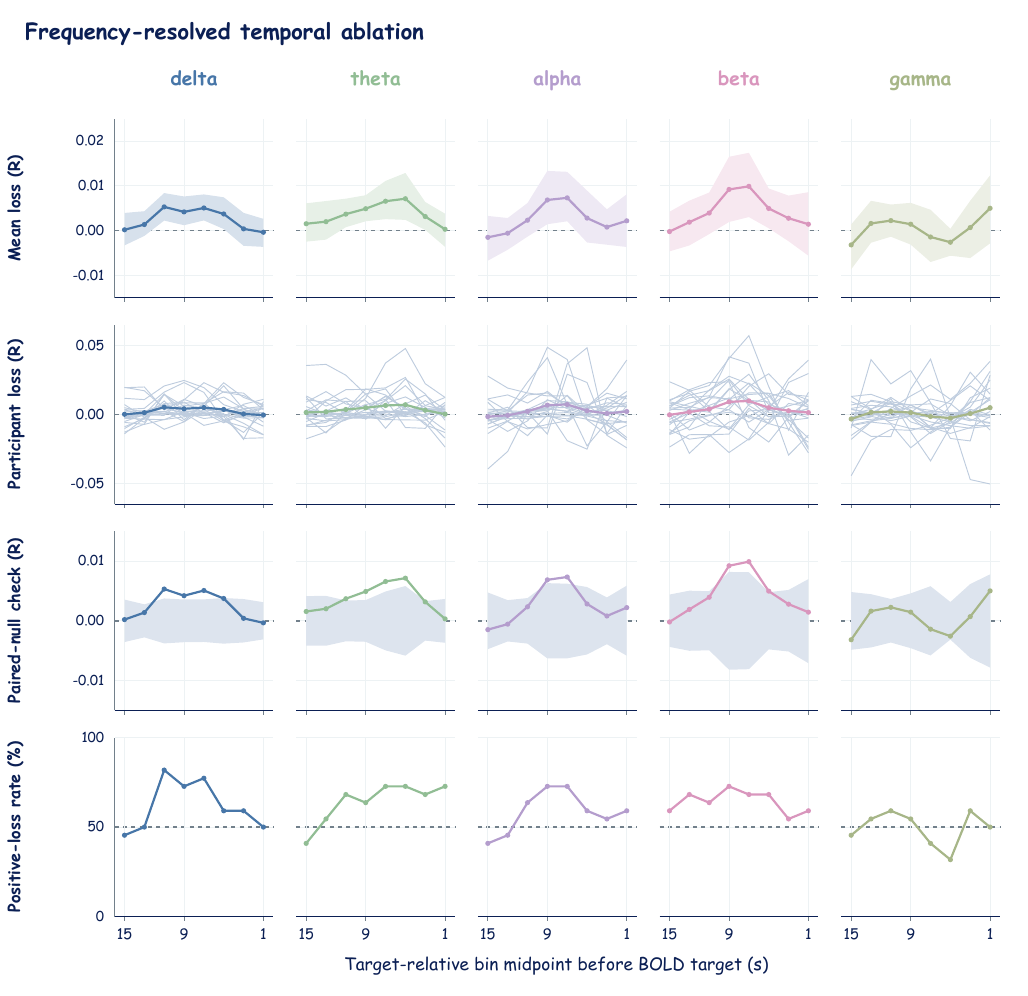}
\caption{Frozen-readout Fourier lag--band sensitivity. Columns denote bands. Rows show participant means with 95\% bootstrap CIs, individual profiles, means against central 95\% paired sign-flip null bands, and positive-loss rates ($n=22$). Loss is full-minus-ablated correlation. Shading is pointwise, and the 50\% line is descriptive.}
\label{fig:appendix_lag_band}
\end{figure*}

\begin{figure*}[pos=tbp]
\centering
\includegraphics[width=\textwidth]{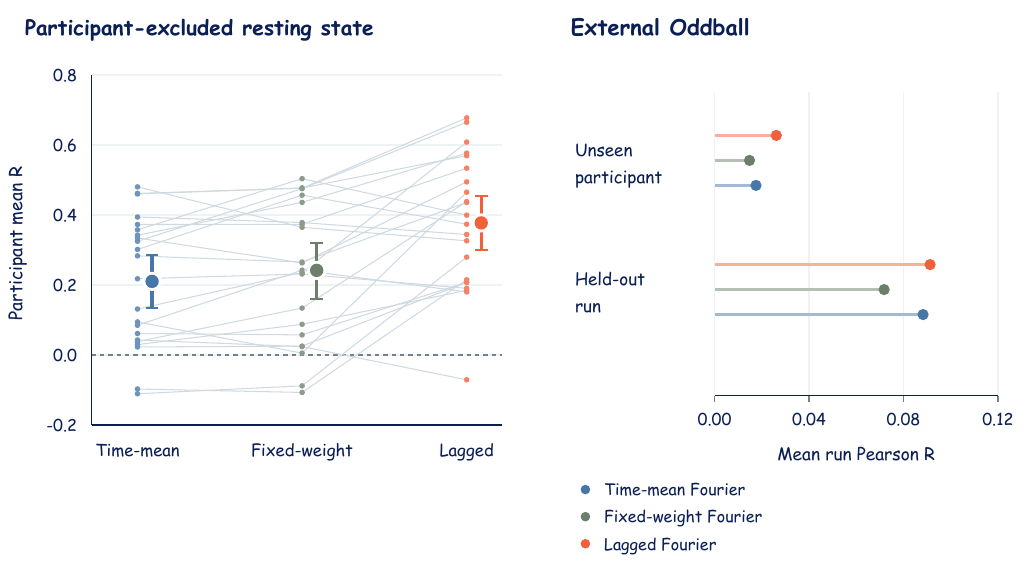}
\caption{Canonical Fourier prediction under distinct generalization protocols. Left: participant-exclusion scores ($n=22$), with means and 95\% bootstrap CIs. Right: means over 102 Oddball runs under participant-grouped and run-key holdouts, without CIs. Models are refitted in both settings.}
\label{fig:appendix_generalization}
\end{figure*}

\section{Coordinate invariance and the scope of disruption controls}
\label{app:coordinates}
Let $P$ be a permutation matrix that reorders the same calibrated features in every observation and partition. Defining $z'_n=Pz_n$ and $w'_r=Pw_r$ gives
\begin{equation}
{z'_n}^{\top}w'_r=z_n^\top w_r,\qquad
\|w'_r\|_2^2=\|w_r\|_2^2.
\label{eq:permutation_invariance}
\end{equation}
Consistent reindexing therefore preserves the Ridge objective, fitted intercept, predictions, and validation criterion up to numerical tolerance. Calibration statistics are permuted with the associated features. This invariance follows from the isotropic penalty; it does not imply that the flat predictor uses spatial or temporal adjacency.

For independent sample-wise permutations, no single coefficient reindexing restores coordinate identity across observations. The resulting loss reflects disruption of the common feature definition together with changes in the distributions presented to the fitted model. Such perturbations assess representation consistency, not a selective intervention on a cortical process.

Circular feature--target shifts are restricted to the designated partitions, while label shuffling changes the training response pairing. Their low correlations demonstrate sensitivity to alignment and supervision under these interventions, but do not define a noise ceiling or a calibrated chance-level test. Every readout in this control family is refitted. A frozen-readout perturbation addresses a different question, considered below.

Figure~\ref{fig:appendix_coordinate_controls} displays the target-wise control outcomes. Shared reindexing produces negligible changes, whereas sample-wise and pairing disruptions affect targets differently. These outcomes are descriptive because participant-level uncertainty was not recorded for this family.

\subsection{Exploratory frozen-readout lag--band sensitivity}
\label{app:lag_band}
The refitted controls assess whether training can recover a consistent measurement-to-response mapping. Figure~\ref{fig:appendix_lag_band} instead examines the response of an already fitted predictor to suppression of part of its test input. This diagnostic uses the historical canonical Fourier cache from the same 29 scans and 22 participants, separately from the matched Morlet experiment. A shared Ridge penalty is selected by mean seven-target validation correlation from $\{0.1,1,10,100,1000,10000\}$, yielding an unperturbed mean scan correlation of approximately 0.452.

Each of 40 interventions sets the cached calibrated coordinates for one frequency band and one two-second interval to zero across channels, before applying the fitted scaler. The scaler and readout remain fixed. Sensitivity is quantified as the full-input minus perturbed-input mean seven-target test correlation. Repeated scans are averaged within participant before computing equal-participant means, CIs, and directional proportions. Because suppression changes the predictor's input distribution and acts on correlated features, the loss does not identify a unique physiological contribution.

Participant-bootstrap CIs use 20,000 resamples (seed 20261013). Null bands use 100,000 paired sign-flip draws (seed 20261012), assuming symmetry of participant effects about zero under the null. Within each draw, one sign per participant is shared across all 40 contrasts, preserving their within-participant dependence. Two-sided Monte Carlo tests use the plus-one correction and Holm adjustment across all 40 cells.

The largest observed mean loss is approximately 0.0099 for the beta-band interval centered seven seconds before the target. No cell passes the 0.05 Holm-adjusted threshold (minimum adjusted $p\approx0.060$). The profiles therefore describe frozen-model sensitivity without statistically resolving a lag-specific effect after multiplicity correction. This post-hoc Fourier analysis uses the previously inspected test partitions and inherited retrospective preprocessing; it is neither independent confirmation of the Morlet gain nor an estimate of a physiological response kernel.

\section{Generalization protocols and illustrative display provenance}
\label{app:generalization}
\subsection{Participant exclusion and external fitting}
The resting-state participant-exclusion analysis begins with uncalibrated log-bandpower tensors. Every scan of the held-out participant is excluded before fitting feature statistics or readout parameters. The remaining participants' training partitions supply the fit and their validation partitions select penalties; evaluation uses the held-out participant's test intervals. This procedure separates participants for feature calibration and prediction fitting while retaining the inherited retrospective BOLD processing.

Additional cached uniform Fourier and whole-history features had already been standardized within each scan before participant exclusion. They are excluded from the main comparison because their calibration does not match the training-participant-only procedure used for canonical Fourier features. The reported 0.377 result consequently refers specifically to canonical lagged Fourier regression, not to Morlet or to every cached representation.

Oddball contains 102 unique runs from 17 participants. Participant-grouped folds exclude all runs from the held-out participant group. Run-key holdouts instead remove one task/run key across participants while retaining their other keys in training. The protocols therefore distinguish unseen participants from unseen runs of represented participants. Correlations are computed within run, averaged over 64 targets, and then averaged over runs.

Oddball readouts are refitted using training-fold feature statistics and fixed penalty 10, with LSQR tolerance $10^{-3}$. Fixed temporal weighting uses a gamma kernel with shape six and scale one, evaluated at reversed lag midpoints and normalized to sum to one; the resting-state comparator instead uses a Gaussian kernel. No fitted resting-state model is transferred unchanged, and external Morlet estimation is not evaluated. Differences in target construction and acquisition further distinguish this endpoint from the seven-target resting-state analysis.

Figure~\ref{fig:appendix_generalization} displays the resting-state and external protocols on separate axes. External means are descriptive because participant-level paired uncertainty is unavailable, and their difference from the resting-state scores is not a matched estimate of transfer loss.

\subsection{EEG traces, scalp interpolation, and coordinate displays}
The illustrative figures use one cached test example from participant 01, scan 01. Short-history traces span $-1.5$ to 0 seconds, are standardized independently within the displayed crop, and are vertically offset. Each scalp snapshot averages approximately $\pm0.12$ seconds around its indicated time before cross-channel standardization. Differences between snapshots therefore describe normalized observations rather than microvolt changes or an event-related average.

In the coordinate figure, the large scalp map shows cross-channel-standardized log variance over approximately $-10$ to $-8$ seconds. The upper sequence shows independently standardized local EEG means. The lower lag--band view uses channel-pooled Welch power with display-specific normalization and a 1--4-Hz delta interval. This descriptive view differs from both prediction estimators; the canonical Fourier delta interval begins at 0.5 Hz.

Scalp interpolation uses recording-electrode positions to visualize the observed EEG, not to estimate cortical sources, readout coefficients, or feature importance. Algorithm tiles and schematic waveforms illustrate computation, whereas paired-score and target-gain figures summarize recorded experimental outcomes. These roles are distinguished so that descriptive input displays are not interpreted as learned biological mechanisms.

\bibliographystyle{cas-model2-names}
\nocite{*}
\bibliography{references}

\end{document}

\let\WriteBookmarks\relax
\shorttitle{MEL supplementary material}
\shortauthors{Liu et al.}
\title[mode=title]{Supplementary material: Target-relative spectral representations for EEG-based regional BOLD prediction}
\author{Xiangyu Liu, Zeting Yan, Zhitong Yin, Boyang Li, and Xi Zhang}
\begin{abstract}
This supplement specifies the analysis families, estimator definitions, participant aggregation, and limitations of the illustrative EEG displays. It reports complete descriptive and paired results for the estimator, matched-readout, and temporal-aggregation experiments. Statistical analyses reaggregate existing scores; they do not retrain models, repair inherited preprocessing, or create an untouched test set.
\end{abstract}
\begin{keywords}
Reproducibility \sep Spectral estimation \sep Participant-level uncertainty
\end{keywords}
\maketitle
\renewcommand{\thetable}{S\arabic{table}}
\renewcommand{\thefigure}{S\arabic{figure}}
\renewcommand{\thesection}{S\arabic{section}}

\section{Estimator specification}
The Morlet centers are 2, 6, 10.5, 21.5, and 37.5 Hz. Cycle counts are 3, 4, 5, 7, and 9. For center $\nu_f$ and cycles $q_f$, the temporal Gaussian standard deviation is $q_f/(2\pi\nu_f)$. The discrete kernel radius is the larger of eight samples and $\lceil3.5f_s q_f/(2\pi\nu_f)\rceil$. Complex kernels are zero-mean and divided by the square root of their summed squared magnitudes, with $10^{-12}$ inside the numerical normalizer. Convolution uses the next power-of-two FFT length at least as long as the history plus kernel minus one. The centered crop returns 3200 samples before averaging eight consecutive 400-sample intervals.

The canonical Fourier representation uses a Hann window independently within each interval. It averages squared real-FFT magnitudes over the five nonoverlapping bands specified in the main text. A uniform-Fourier representation changes only the frequency intervals to five equally spaced ranges between 0.5 and 45 Hz. Whole-history spectral features use a single Hann-windowed history and 40 frequency intervals per channel. Although this comparator is called ``PSD'' in some analysis records, its implementation is not a density-normalized physical PSD. All three equal-dimensional coordinate representations and the whole-history comparator contain 1040 scalar features.

\section{Distinct analysis families}
The matched-readout experiment uses a 33-value Ridge grid from $10^{-1}$ to $10^7$ and a shared 16-hidden-unit MLP with five seeds. An individual MLP has 18,855 trainable parameters for a 1040-dimensional input, including layer normalization. A five-model prediction ensemble contains five fitted MLPs; it is not a single-seed result. Canonical component-ablation values are drawn from the corresponding 33-penalty family.

The estimator-and-aggregation experiment uses a 31-value grid from $10^{-2}$ to $10^8$ and LSQR Ridge. Pooling is performed on calibrated coordinates, followed by training-only scaling of the resulting feature vector. Dimensions are 40 for channel pooling, 208 for frequency pooling, and 130 for lag pooling. Axis marginals concatenate lag--frequency, channel--frequency, and lag--channel averages (378 features); means and standard deviations over channels give 80 features. PCA and PLS are fitted to training inputs. The validation-chosen PCA setting is 64 components without whitening; PLS uses four components.

The broader validation experiment also evaluates global and ROI-wise combinations of canonical, uniform-Fourier, full Morlet, and channel-pooled Morlet predictions. ROI-wise fusion and axis marginals have higher mean validation scores than full Morlet features. Accordingly, the manuscript reports Morlet as the strong simple estimator in the controlled comparison, not as the globally validation-optimal candidate across all evaluated representations. No post-hoc test ranking is used to relabel this selection history.

The temporal experiment uses a separate 17-value grid from $10^{-3}$ to $10^5$. Fixed Gaussian weighting centers the eight lag midpoints at a seven-second delay with width 2.5 seconds. Its derivative comparator includes the temporal derivative of this kernel. The smoothing comparator uses weights $(0.25,0.5,0.25)$ with edge padding. These are descriptive spectral regression controls, not fitted physiological hemodynamic response functions.

\section{Participant aggregation and paired uncertainty}
For each scan, target-wise Pearson correlations are averaged arithmetically. Scans are then averaged within participant. Thus, participants with two scans do not receive twice the weight of participants with one scan. Bootstrap samples resample 22 participant averages with replacement 20,000 times using seed 20261011. The interval for a paired difference resamples participant differences directly, retaining pairing between methods. Bootstrap percentile intervals are not multiplicity-adjusted.

An exact two-sided sign test counts positive and negative participant differences, excluding exact zeros. Holm correction is applied to every comparison with the reference method in each reported family: 14 in the estimator experiment, 11 in the matched-readout experiment, seven in the temporal experiment, and two in the canonical participant-exclusion comparison. These family definitions do not account for every historical design choice made during development. Previously inspected test sets and upstream whole-scan preprocessing therefore remain limitations despite the corrected within-family comparisons.

% BEGIN EMBEDDED tables_supplementary.tex
\begin{table*}[pos=p]
\centering\scriptsize
\setlength{\tabcolsep}{4pt}
\renewcommand{\arraystretch}{1.15}
\begin{tabular}{p{0.46\textwidth}rrrr}
\toprule
Representation/readout & Validation & Scan mean & Participant mean & 95\% interval \\
\midrule
Canonical lagged bandpower & 0.478 & 0.431 & 0.436 & [0.370, 0.503] \\
Classical HRF-weighted bandpower & 0.428 & 0.312 & 0.317 & [0.238, 0.390] \\
Classical time-mean bandpower & 0.367 & 0.258 & 0.262 & [0.189, 0.328] \\
Flattened PSD & 0.485 & 0.363 & 0.368 & [0.285, 0.447] \\
MEL fusion, ROI-wise weights & 0.558 & 0.506 & 0.512 & [0.450, 0.573] \\
MEL fusion, global weights & 0.552 & 0.501 & 0.507 & [0.443, 0.570] \\
Morlet PCA-Ridge d=64 whiten=0 & 0.473 & 0.497 & 0.498 & [0.433, 0.560] \\
Morlet PLS components=4 & 0.458 & 0.486 & 0.488 & [0.406, 0.564] \\
Morlet axis marginals & 0.558 & 0.460 & 0.455 & [0.389, 0.518] \\
Morlet channel-band, lag pooled & 0.441 & 0.339 & 0.352 & [0.287, 0.412] \\
Morlet lag-band mean+std & 0.500 & 0.471 & 0.478 & [0.416, 0.541] \\
Morlet lag-band, channel pooled & 0.515 & 0.481 & 0.480 & [0.419, 0.542] \\
Morlet lag-channel, band pooled & 0.493 & 0.426 & 0.421 & [0.363, 0.481] \\
Morlet lag-channel-band & 0.543 & 0.507 & 0.505 & [0.443, 0.565] \\
Uniform STFT lag-band & 0.492 & 0.455 & 0.464 & [0.399, 0.529] \\
\bottomrule\end{tabular}
\caption{Estimator and aggregation experiment: 31 penalties and LSQR Ridge. Intervals resample participants after within-participant scan averaging.}
\end{table*}
\begin{table*}[pos=p]
\centering\scriptsize
\setlength{\tabcolsep}{4pt}
\renewcommand{\arraystretch}{1.15}
\begin{tabular}{p{0.46\textwidth}rrrr}
\toprule
Comparator & Mean difference & 95\% interval & Wins/22 & Adjusted $p$ \\
\midrule
Canonical lagged bandpower & +0.069 & [+0.037, +0.101] & 19/22 & 0.0086 \\
Classical HRF-weighted bandpower & +0.189 & [+0.139, +0.238] & 21/22 & 0.0002 \\
Classical time-mean bandpower & +0.243 & [+0.177, +0.312] & 21/22 & 0.0002 \\
Flattened PSD & +0.137 & [+0.087, +0.196] & 20/22 & 0.0015 \\
MEL fusion, ROI-wise weights & -0.007 & [-0.024, +0.012] & 9/22 & 1.0000 \\
MEL fusion, global weights & -0.001 & [-0.023, +0.020] & 8/22 & 1.0000 \\
Morlet PCA-Ridge d=64 whiten=0 & +0.007 & [-0.022, +0.035] & 13/22 & 1.0000 \\
Morlet PLS components=4 & +0.017 & [-0.028, +0.071] & 9/22 & 1.0000 \\
Morlet axis marginals & +0.050 & [+0.017, +0.091] & 17/22 & 0.1352 \\
Morlet channel-band, lag pooled & +0.153 & [+0.078, +0.225] & 18/22 & 0.0391 \\
Morlet lag-band mean+std & +0.027 & [+0.001, +0.052] & 16/22 & 0.3674 \\
Morlet lag-band, channel pooled & +0.025 & [+0.004, +0.047] & 14/22 & 1.0000 \\
Morlet lag-channel, band pooled & +0.084 & [+0.048, +0.119] & 20/22 & 0.0015 \\
Uniform STFT lag-band & +0.041 & [+0.004, +0.077] & 14/22 & 1.0000 \\
\bottomrule\end{tabular}
\caption{Paired comparisons against Morlet lag-channel-band. Positive differences favor the reference. Holm correction covers all 14 comparisons in this experiment. Bootstrap intervals are unadjusted.}
\end{table*}
\begin{table*}[pos=p]
\centering\scriptsize
\setlength{\tabcolsep}{4pt}
\renewcommand{\arraystretch}{1.15}
\begin{tabular}{p{0.46\textwidth}rrrr}
\toprule
Representation/readout & Validation & Scan mean & Participant mean & 95\% interval \\
\midrule
Classical lagged bandpower / MEL-flat | Matched MLP & 0.389 & 0.374 & 0.369 & [0.306, 0.433] \\
Classical lagged bandpower / MEL-flat | Matched MLP Ensemble & 0.438 & 0.431 & 0.426 & [0.359, 0.492] \\
Classical lagged bandpower / MEL-flat | Ridge & 0.478 & 0.432 & 0.437 & [0.370, 0.504] \\
Flattened PSD | Matched MLP & 0.339 & 0.302 & 0.304 & [0.236, 0.372] \\
Flattened PSD | Matched MLP Ensemble & 0.394 & 0.356 & 0.361 & [0.285, 0.434] \\
Flattened PSD | Ridge & 0.485 & 0.363 & 0.368 & [0.285, 0.447] \\
Morlet wavelet energy | Matched MLP & 0.445 & 0.433 & 0.428 & [0.357, 0.495] \\
Morlet wavelet energy | Matched MLP Ensemble & 0.496 & 0.491 & 0.487 & [0.412, 0.556] \\
Morlet wavelet energy | Ridge & 0.543 & 0.508 & 0.506 & [0.444, 0.565] \\
Uniform STFT bins | Matched MLP & 0.402 & 0.368 & 0.368 & [0.310, 0.428] \\
Uniform STFT bins | Matched MLP Ensemble & 0.459 & 0.429 & 0.430 & [0.367, 0.492] \\
Uniform STFT bins | Ridge & 0.492 & 0.456 & 0.465 & [0.400, 0.529] \\
\bottomrule\end{tabular}
\caption{Matched-readout experiment: 33-penalty Ridge and fixed MLP architecture. Intervals resample participants after within-participant scan averaging.}
\end{table*}
\begin{table*}[pos=p]
\centering\scriptsize
\setlength{\tabcolsep}{4pt}
\renewcommand{\arraystretch}{1.15}
\begin{tabular}{p{0.46\textwidth}rrrr}
\toprule
Comparator & Mean difference & 95\% interval & Wins/22 & Adjusted $p$ \\
\midrule
Classical lagged bandpower / MEL-flat | Matched MLP & +0.136 & [+0.095, +0.180] & 21/22 & 0.0001 \\
Classical lagged bandpower / MEL-flat | Matched MLP Ensemble & +0.080 & [+0.039, +0.123] & 15/22 & 0.4014 \\
Classical lagged bandpower / MEL-flat | Ridge & +0.069 & [+0.037, +0.101] & 19/22 & 0.0060 \\
Flattened PSD | Matched MLP & +0.201 & [+0.153, +0.249] & 20/22 & 0.0011 \\
Flattened PSD | Matched MLP Ensemble & +0.145 & [+0.093, +0.196] & 19/22 & 0.0060 \\
Flattened PSD | Ridge & +0.138 & [+0.088, +0.196] & 20/22 & 0.0011 \\
Morlet wavelet energy | Matched MLP & +0.077 & [+0.041, +0.125] & 19/22 & 0.0060 \\
Morlet wavelet energy | Matched MLP Ensemble & +0.019 & [-0.022, +0.069] & 12/22 & 0.8318 \\
Uniform STFT bins | Matched MLP & +0.137 & [+0.093, +0.179] & 21/22 & 0.0001 \\
Uniform STFT bins | Matched MLP Ensemble & +0.076 & [+0.033, +0.117] & 19/22 & 0.0060 \\
Uniform STFT bins | Ridge & +0.041 & [+0.005, +0.076] & 14/22 & 0.5726 \\
\bottomrule\end{tabular}
\caption{Paired comparisons against Morlet wavelet energy | Ridge. Positive differences favor the reference. Holm correction covers all 11 comparisons in this experiment. Bootstrap intervals are unadjusted.}
\end{table*}
\begin{table*}[pos=p]
\centering\scriptsize
\setlength{\tabcolsep}{4pt}
\renewcommand{\arraystretch}{1.15}
\begin{tabular}{p{0.46\textwidth}rrrr}
\toprule
Representation/readout & Validation & Scan mean & Participant mean & 95\% interval \\
\midrule
Canonical HRF BP & 0.425 & 0.322 & 0.328 & [0.248, 0.403] \\
Distributed lag-band BP & 0.449 & 0.449 & 0.455 & [0.395, 0.515] \\
Distributed lag-channel BP & 0.459 & 0.393 & 0.397 & [0.339, 0.456] \\
FIR lag-only BP & 0.390 & 0.379 & 0.386 & [0.321, 0.448] \\
Full MEL & 0.475 & 0.431 & 0.436 & [0.369, 0.504] \\
HRF + derivative BP & 0.493 & 0.424 & 0.430 & [0.370, 0.491] \\
Temporal smoothing regression & 0.483 & 0.430 & 0.435 & [0.371, 0.499] \\
Time-mean BP & 0.269 & 0.302 & 0.314 & [0.250, 0.376] \\
\bottomrule\end{tabular}
\caption{Canonical-Fourier temporal experiment: 17 validation-selected penalties. Intervals resample participants after within-participant scan averaging.}
\end{table*}
\begin{table*}[pos=p]
\centering\scriptsize
\setlength{\tabcolsep}{4pt}
\renewcommand{\arraystretch}{1.15}
\begin{tabular}{p{0.46\textwidth}rrrr}
\toprule
Comparator & Mean difference & 95\% interval & Wins/22 & Adjusted $p$ \\
\midrule
Canonical HRF BP & +0.108 & [+0.061, +0.155] & 18/22 & 0.0304 \\
Distributed lag-band BP & -0.019 & [-0.047, +0.006] & 8/22 & 0.8588 \\
Distributed lag-channel BP & +0.039 & [+0.005, +0.071] & 15/22 & 0.5352 \\
FIR lag-only BP & +0.050 & [+0.013, +0.087] & 16/22 & 0.3149 \\
HRF + derivative BP & +0.007 & [-0.023, +0.037] & 13/22 & 1.0000 \\
Temporal smoothing regression & +0.001 & [-0.011, +0.014] & 12/22 & 1.0000 \\
Time-mean BP & +0.122 & [+0.060, +0.180] & 16/22 & 0.3149 \\
\bottomrule\end{tabular}
\caption{Paired comparisons against Full MEL. Positive differences favor the reference. Holm correction covers all 7 comparisons in this experiment. Bootstrap intervals are unadjusted.}
\end{table*}
% END EMBEDDED tables_supplementary.tex

\section{Participant exclusion and external protocols}
The canonical participant-exclusion experiment uses raw log-bandpower tensors, not features calibrated within the held-out scan. Scalers and models are fitted to training partitions from the other participants, and their validation partitions determine each target's penalty. All scans of the held-out participant are removed from fitting and model selection. The inherited BOLD filtering and whole-scan target normalization remain in place. Results assess unseen-participant regression under that offline preprocessing, not prospective deployment.

The complete cached traditional-comparator experiment additionally contains uniform-Fourier and whole-history spectral rows. Those feature caches were already standardized within each scan before the participant-exclusion experiment. They are omitted from the main participant-exclusion table because they do not share the canonical representation's strictly training-participant feature calibration. Their slightly different scores are not used to assert a better unseen-participant representation.

Oddball folds refit canonical spectral regressions on the external data. The fixed temporal weighting is a gamma kernel with shape six and scale one, evaluated at reversed lag midpoints and normalized to sum to one. This differs from the Gaussian comparator used for resting-state data. Run-key holdouts remove one task/run key across all participants, leaving the other keys for training; these participants are represented in training. The participant-grouped experiment excludes all runs of a held-out participant group. Neither experiment transfers a fitted resting-state model unchanged.

\section{Additional temporal and cache checks}
The temporal design analysis compares one-, two-, four-, and eight-second Fourier bins under a historical 33-penalty protocol with a shared penalty selected over all targets. Their mean scan-wise scores are approximately 0.447, 0.452, 0.430, and 0.409, respectively. The shared-penalty two-second result of 0.452 differs from the current target-specific-penalty result of 0.432. These values are reported as sensitivity analysis, not replacements for the matched-readout table's headline values.

The cache-gap experiment gives mean scan-wise scores of 0.431, 0.433, 0.432, 0.429, and 0.425 for extra guards of 0, 4, 8, 16, and 24 target times. This checks removal of neighboring cached observations. It does not rerun raw filtering or whole-scan normalization and cannot establish that those operations are split-safe.

The sensitivity displays are included in the main article. The independently sampled training-fraction subsets cannot support a paired comparison of sample efficiency between representations. This supplement retains their configuration details without duplicating the figures.

\section{Illustrative display definitions}
The EEG and coordinate displays in the main article use one cached test example (participant 01, scan 01). They do not depict a participant-averaged neural mechanism, an ERP, a cortical source estimate, or learned feature importance. Display normalization is separate from the training-derived feature calibration used for prediction.

In the short-history display, traces are independently standardized within the $-1.5$- to 0-second crop and vertically offset. Scalp snapshots average approximately $\pm0.12$ seconds around each indicated time and are standardized across channels. Heights are not absolute microvolt amplitudes.

In the coordinate display, the left map shows cross-channel-standardized log variance over approximately $-10$ to $-8$ seconds. Upper maps show independently standardized local mean EEG. The lower view uses channel-pooled Welch power with a 1--4-Hz delta interval and display-level normalization. This is not the exact Morlet input or canonical Fourier representation; the latter begins its delta interval at 0.5 Hz. Interpolation uses recorded electrode positions rather than cortical source reconstruction.

The canonical token-computation diagram is paired with the estimator comparison in the main article. Its Hann-windowed real-FFT branch is distinguished from the Morlet history-convolution branch; neither diagram represents a fitted importance map. Figures are not repeated in this supplement.